\documentclass{article}

\usepackage[preprint]{corl_2026} 

\usepackage{graphicx}
\usepackage{amsmath}
\usepackage{amssymb}
\usepackage{array}
\usepackage{tabularx}
\usepackage{multirow}
\usepackage[table]{xcolor}  
\usepackage{booktabs}
\usepackage{makecell}
\usepackage{caption}
\usepackage{subcaption}    
\usepackage{algorithm}
\usepackage{algpseudocode}
\usepackage{soul}         
\usepackage{wrapfig}

\title{Blind Dexterous Grasping via Real2Sim2Real Tactile Policy Learning}
\author{
  Shengcheng Luo$^{1,2,*}$, Xiyan Huang$^{1,*}$, Zhe Xu$^{1}$, Wanlin Li$^{2}$, Ziyuan Jiao$^{2,\dagger}$, Chenxi Xiao$^{1,\dagger}$ \\
  $^{1}$ShanghaiTech University \hspace{0.4cm}
  $^{2}$Beijing Institute for General Artificial Intelligence \\
  $^{*}$Equal contribution. \quad $^{\dagger}$Corresponding authors.
}

\begin{document}
\maketitle

\vspace{-30px}

\begin{center}
\includegraphics[width=\textwidth]{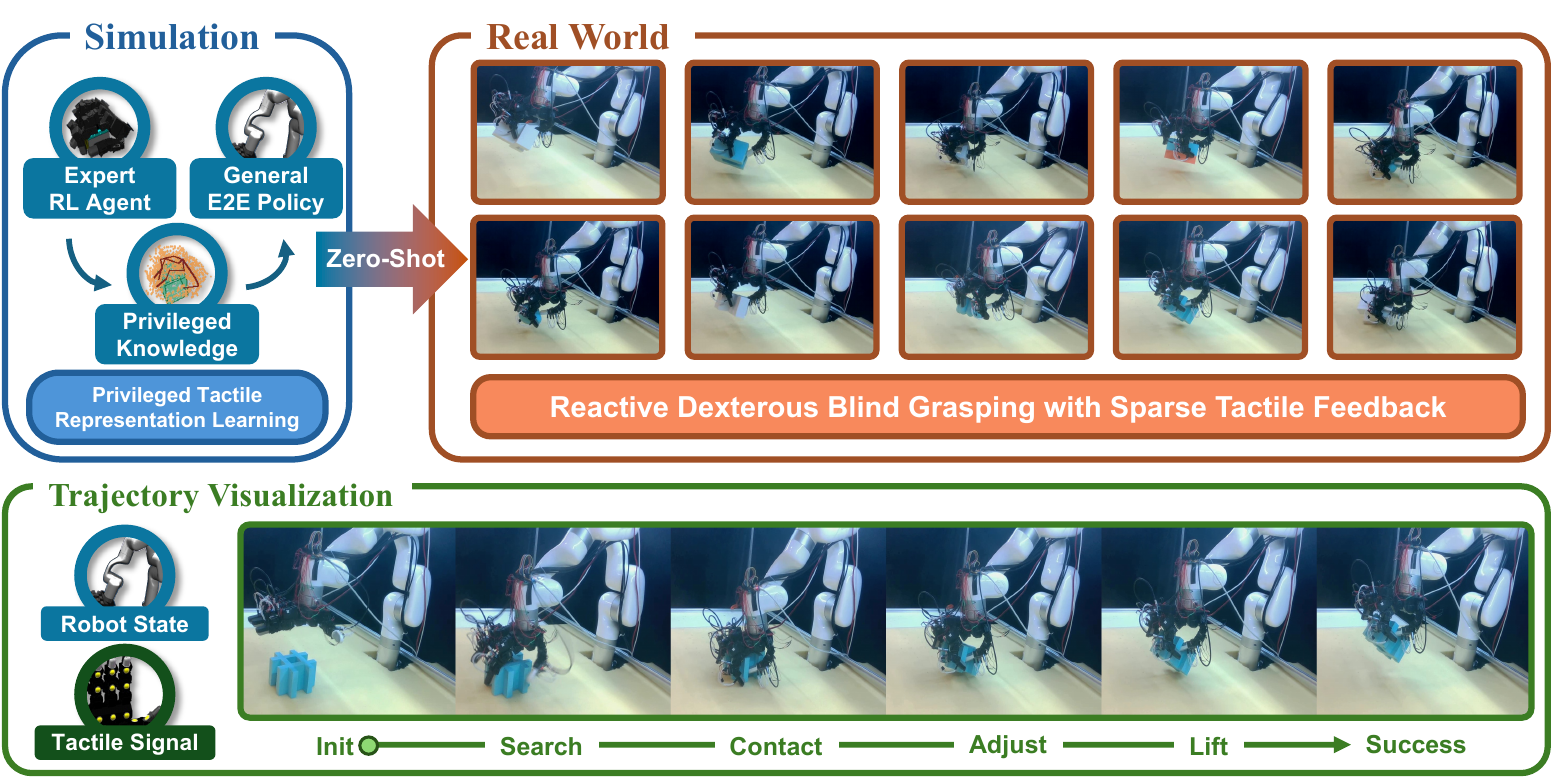}
\captionof{figure}{\textbf{Zero-shot sim-to-real transfer of a tactile-conditioned control policy for reactive blind grasping.} The sequence demonstrates the deployed policy's ability to perform blind object interaction, including dynamically searching for the object, adjusting grasp contacts, and lifting, all without visual input and using only robot state and sparse tactile feedback.}
\label{fig:teaser}
\end{center}

\vspace{-10px}


\begin{abstract}
Blind grasping with a dexterous hand is a crucial manipulation capability. Nevertheless, learning such tactile-only policies for real robots remains challenging due to the tactile sim-to-real gap and the limited expressiveness of sparse tactile signals.
To bridge this gap, we propose a framework for tactile-only blind grasping that is deployable on a physical multi-fingered robotic hand.
Our approach combines three key components. First, we introduce a Real2Sim tactile calibration pipeline that constructs a contact-calibrated digital-twin simulator capable of reproducing real tactile signals.
Second, we improve the expressiveness of sparse tactile observations using a layout-aware tactile encoder, which incorporates sensor-geometry priors through self-supervised pretraining.
Third, to improve generalization to unseen objects, we train object-specific reinforcement-learning experts in the calibrated simulator and aggregate their successful grasp trajectories into a tactile-conditioned Diffusion Policy.
We evaluate our method on a physical LEAP Hand equipped with distributed tactile sensing across 10 seen and 10 unseen objects.
The deployed policy achieves a 27\% real-world grasp success rate across all 20 objects, without real-world grasping demonstrations or visual input.
Simulation ablations show that layout-aware tactile pretraining improves grasping performance, while sensing-level evaluations confirm that Real2Sim calibration increases the consistency of tactile contact events between simulation and hardware.
Together, these results suggest that contact-event calibration, geometry-aware tactile representation learning, and diffusion-based policy aggregation provide an effective path toward tactile-only blind grasping on real dexterous robotic hands. Project page: \href{https://Dex-Blind-Grasp.github.io/}{Dex-Blind-Grasp.github.io}.
\end{abstract}



\section{Introduction}
Blind grasping is essential for human and robot manipulation when vision is unreliable, occluded, or unavailable~\cite{2018More,2018The,yin2023rotatingseeinginhanddexterity}. Rather than relying on visual estimates of object pose or geometry, blind grasping uses contact feedback to form stable grasps through reactive physical interaction. It is therefore well suited to darkness, clutter, narrow spaces, and severe occlusion, where camera-based perception often fails~\cite{conf/icra/DangWA11}. This makes tactile-only blind grasping a promising capability for dexterous manipulation under visual uncertainty.

Despite these advantages, achieving tactile-only blind grasping on physical robotic hands remains challenging. Demonstration-based learning via teleoperation is difficult to scale, as dexterous in-hand object search and adjustment under blind conditions require substantial human situational awareness and effort. Simulation and reinforcement learning offer scalable alternatives, but tactile policies remain difficult to transfer to real hardware for two main reasons. First, tactile sensing is highly sensitive to material properties, sensor placement, manufacturing variation, and contact mechanics, all of which are difficult to model accurately~\cite{10.1016/j.robot.2015.07.015,church2021tactile}. Second, sparse tactile arrays provide limited geometric information and only partial observations of the hand--object state. The same tactile reading may correspond to many object poses, contact configurations, and grasping motions~\cite{garciagarcia2019tactilegcngraphconvolutionalnetwork}. Thus, tactile-only grasping requires both well-calibrated contact simulation and policies that can reason under partial tactile observability.

In this work, we propose a hardware-deployable framework for tactile-only blind grasping with a physical dexterous hand. Our first goal is to build a digital-twin environment with a minimal tactile sim-to-real gap. Prior work has used binary tactile signals to reduce domain shift~\cite{li2025taccel,si2021taximexamplebasedsimulationmodel}, showing that contact-event abstractions are more transferable than high-dimensional tactile measurements. However, we find that even binary tactile signals exhibit a non-negligible sim-to-real discrepancy, mainly due to differences in contact onset and offset timing. To address this, we introduce a lightweight Real2Sim tactile calibration procedure that aligns simulated and real contact events by tuning contact margins and activation thresholds. This produces a calibrated digital-twin simulator that reproduces real tactile contact timing without complex soft-body modeling or real-world grasping demonstrations.

Our second goal is to improve the expressiveness of sparse tactile observations. Since raw binary contacts provide limited information about object geometry and hand--object configuration, we augment the policy with a layout-aware encoder that incorporates each sensor's 3D location on the hand. This enables the policy to reason about where contacts occur. We further pretrain the encoder in simulation using privileged geometric supervision, including object shape, hand--object contact labels, and object pose. This encourages the latent representation to capture object- and contact-level structure that is not directly observable from sparse tactile inputs alone.

Finally, to handle diverse object geometries and multimodal grasping strategies, we combine reinforcement learning with diffusion-based policy learning. We first train object-specific RL experts in the calibrated simulator to discover successful tactile grasping behaviors. We then aggregate their successful trajectories and distill them into a tactile-conditioned Diffusion Policy~\cite{chi2024diffusionpolicyvisuomotorpolicy}. This enables the deployed controller to generate multiple valid grasping motions conditioned on tactile history, improving robustness when grasping unseen objects under partial observability.

We evaluate our framework on a physical LEAP Hand equipped with distributed tactile sensing. Across 20 diverse objects, including 10 seen and 10 unseen instances, our deployed policy achieves \textbf{27\%} grasp success. These results demonstrate the feasibility of transferring tactile-only blind grasping policies to real hardware using contact-event calibration, layout-aware tactile representation learning, and tactile-conditioned diffusion policy learning.

Our main contributions are summarized as follows:
\begin{itemize}
    \item We propose a \textit{Real2Sim2Real} framework for tactile-only blind dexterous grasping that enables deployment on a physical robotic hand without real-world grasping demonstrations.
    \item We introduce a lightweight contact-event calibration method that reduces the binary tactile sim-to-real gap by aligning contact onset and offset timing between simulation and hardware.
    \item We develop a layout-aware tactile encoder that exploits the spatial layout of tactile sensors and uses privileged pretraining to improve reasoning from sparse tactile observations.
    \item We introduce an expert-to-diffusion policy learning scheme that distills diverse object-specific RL grasping behaviors into a unified tactile-conditioned Diffusion Policy, enabling multimodal grasp generation under partial tactile observability.
\end{itemize}


\section{Related Works}
\textbf{Tactile-Driven Dexterous Manipulation.}
Tactile sensing provides direct contact information that is difficult to obtain from vision alone, especially under occlusion, self-occlusion, and uncertain contact conditions. It has therefore become an important modality for dexterous manipulation, supporting grasp stability estimation~\cite{2018More,2018The}, slip detection~\cite{hu2024learningdetectsliptactile,509205}, in-hand manipulation~\cite{lin2024twisting}, and contact-rich dexterous control~\cite{yuan2024robotsynesthesiainhandmanipulation,lin2025pp}. Recent high-resolution fingertip sensors further extend tactile perception beyond binary or force-level feedback, enabling richer local contact understanding such as contact geometry reconstruction~\cite{huang2026gelslamrealtimehighfidelityrobust}, shear estimation~\cite{7139016}, and object pose estimation~\cite{bauza2020tactileobjectposeestimation,8246881}. Despite these advances, most existing work still relies on visuotactile fusion, while vision-free dexterous grasping remains relatively underexplored.

\textbf{Tactile Calibration and Sim-to-Real Transfer.}
For tactile policies, the sim-to-real gap remains a major challenge due to the limited fidelity of soft-material and contact simulation. Existing calibration methods typically identify physical parameters for specific sensors and use them in simulation~\cite{ding2021simtorealtransferroboticmanipulation,Lach2023TowardsTT,Chen2024GeneralPurposeSP}. However, scaling such continuous system identification to dozens of heterogeneous sensors across a dexterous hand is difficult and has been only minimally explored~\cite{narang2021simtorealrobotictactilesensing,Church2021TactileSP}. Recent work has instead explored domain-invariant sensory abstractions, such as binary tactile signals, which have proven effective for sim-to-real transfer~\cite{Lee_2024,touch-dexterity}. However, even binary event streams can exhibit temporal mismatches between simulation and the real world, an issue that remains largely unaddressed. To address this gap, we present a lightweight Real2Sim calibration pipeline for full-hand binary tactile event streams, enabling efficient calibration for zero-shot sim-to-real transfer.

\textbf{Geometric Priors in Tactile Representation.}
Dexterous grasping requires reasoning about object geometry and contact layout~\cite{Liu_2022}. Vision-based policies can access global shape from visual observations~\cite{chen2025bodexscalableefficientrobotic,chen2025dexonomy,fei2025trograspefficientgraph}, whereas tactile-only methods often encode sparse activations directly~\cite{yang2024anyrotategravityinvariantinhandobject,Lee_2024}, making their spatial context on an articulated hand ambiguous. Prior work has shown that tactile feedback can infer object shape and pose~\cite{lee2025vitascopevisuotactileimplicitrepresentation,bauza2020tactileobjectposeestimation,10246361}, and recent methods spatially anchor high-resolution visuotactile signals using hand kinematics~\cite{huang2026spatiallyanchoredtactileawareness,huang20253dvitaclearningfinegrainedmanipulation}. However, these approaches rely on locally dense fingertip tactile readings and are not directly applicable to sparse binary sensors distributed across a full hand. To fill this gap, we introduce a layout-aware tactile encoder pretrained with privileged simulator supervision, injecting object geometry and dense contact priors into sparse full-hand tactile observations.


\section{Learning Tactile Blind Grasping}
\begin{figure*}[t]
    \centering
    \includegraphics[width=\textwidth]{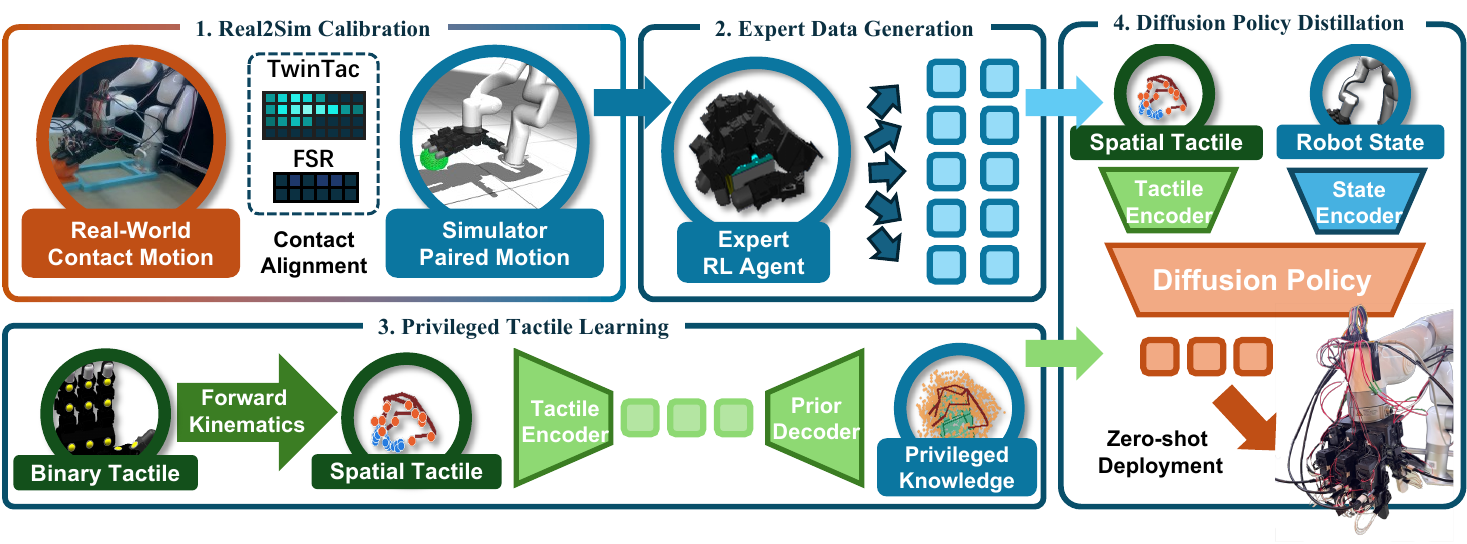}
    \caption{Overview of the proposed Real2Sim2Real framework for blind dexterous grasping. We calibrate tactile events in a paired simulator, generate simulated grasping demonstrations with privileged labels, pretrain a layout-aware tactile encoder, and train a tactile-conditioned Diffusion Policy for real-world deployment.}
    \label{fig:system_architecture}
    \vspace{-10px}
\end{figure*}

An overview of our blind grasping method is shown in Fig.~\ref{fig:system_architecture}. The goal is to stably lift objects with a physical multi-fingered dexterous hand using only proprioceptive and tactile feedback, without visual observations, object pose estimates, or object geometry.

To achieve the goal, our proposed framework consists of three stages. First, we construct a paired simulator with the same nominal robot kinematics and tactile sensor layout as the hardware, and calibrate its binary tactile events to match the contact-onset behavior of the physical sensors (Sec.~\ref{subsec:hardware}). Second, we pretrain a layout-aware tactile encoder using simulation-only geometric and contact supervision from the collected dataset (Sec.~\ref{subsec:representation}). Third, we generate an offline dataset of successful tactile-proprioceptive grasping trajectories in the calibrated simulator and train a tactile-conditioned Diffusion Policy that can be deployed on the physical hand (Sec.~\ref{subsec:policy}).

\subsection{Hardware and Paired Simulator}
\label{subsec:hardware}
\paragraph{Hardware Setup.} We first describe the hardware setup used for blind grasping. Our physical platform consists of a 6-DoF xArm6 manipulator equipped with a 16-DoF LEAP Hand~\cite{shaw2023leaphandlowcostefficient}. The hand is instrumented with distributed tactile sensing across the fingertips, phalanges, and palm. Specifically, we mount four custom-built TwinTac sensors~\cite{Xiyan2025Twintac} on the fingertips and 12 commercial force-sensing resistors (FSRs) on the palm and finger surfaces. Each TwinTac sensor provides eight binary contact channels, and each FSR provides one binary channel, yielding $N_{\mathrm{tac}}=44$ tactile channels in total. The robot receives proprioceptive feedback from its actuated joints and binary tactile activations from these heterogeneous sensors. Additional details are provided in Appendix~\ref{app:hardware}.

\paragraph{Contact Modeling.}
Following prior work~\cite{Lee_2024}, our policy uses binary contact events to simplify sim-to-real transfer. However, differences in sensor sensitivity, material properties, and contact mechanics can cause the onset and offset of binary contact activations to differ substantially between simulation and real hardware. To reduce this discrepancy, we introduce a tactile contact model in our simulation environment.

Specifically, we discretize each sensing region into 3D surface nodes and query their proximity to the object using signed distance fields (SDFs)~\cite{KaolinLibrary}. For the FSR sensors on the palm and phalanges, which require a minimum actuation force to register contact, we approximate activation using a calibrated spatial threshold. An FSR is marked as active if any node in its sensing region has an SDF value below $\lambda_{\mathrm{con}}$.

For the fingertip TwinTac sensors, contact-induced deformation can propagate through the soft elastomer layer, causing adjacent taxels to respond even without direct geometric intersection~\cite{vinogradova2003interaction}. To model this cross-talk, we first compute the raw contact pressure $f_j$ at each simulated surface node $\mathbf{p}_j$ using a penalty-based model~\cite{xu2022efficient}. We then define virtual taxels $T=\{\mathbf{t}_1,\mathbf{t}_2,\ldots,\mathbf{t}_8\}$ according to the hardware layout. The propagated pressure at the $i$-th taxel is computed as
\begin{equation}
    f^t_i = \sum_{j=1}^{n} \exp\left(-\alpha \left\|\mathbf{p}_j-\mathbf{t}_i\right\|\right) f_j ,
\end{equation}
where $\alpha$ controls the spatial decay rate. The $i$-th taxel is activated if $f^t_i > \lambda_{\mathrm{pre}}$. This spatial aggregation reduces sensitivity to minor mesh penetrations while preserving correlated multi-taxel activation patterns.

\paragraph{Real-to-Sim Tactile Sensor Calibration.}
Next, we calibrate the simulator at the level of binary tactile events. Let $\theta=(\lambda_{\mathrm{con}},\alpha,\lambda_{\mathrm{pre}})$ denote the tactile simulation parameters. We collect calibration data by executing task-agnostic primitive motions on the real hand, including tapping and sliding against a known surface, and record synchronized joint trajectories and binary tactile readings $\mathbf{y}^{\mathrm{real}}_{1:T}$. We then replay the same joint trajectories in simulation to obtain the corresponding simulated tactile readings $\mathbf{y}^{\mathrm{sim}}_{1:T}(\theta)$. The parameters are estimated by minimizing the $\ell_1$ mismatch between the real and simulated binary event sequences:
\begin{equation}
    \theta^\star
    =
    \arg\min_{\theta}
    \sum_{t=1}^{T}
    \left\|
    \mathbf{y}^{\mathrm{sim}}_{t}(\theta)
    -
    \mathbf{y}^{\mathrm{real}}_{t}
    \right\|_1 .
\end{equation}
We solve this low-dimensional optimization using grid search over bounded ranges of $\theta$.

\paragraph{Simulation Demonstration Data.}
After tactile calibration, we use the calibrated simulator to collect successful grasping trajectories. We train object-specific PPO experts solely as data-generation policies. The expert actors receive only deployable observations, namely proprioception and binary tactile activations. During training, we use an asymmetric actor-critic setup, where the critic and reward computation access privileged simulator quantities to improve sample efficiency.
After convergence, we roll out the expert actors from randomized initial object placements and retain successful grasp-and-lift trajectories. This yields an offline dataset $\mathcal{D}$ containing observation-action trajectories together with synchronized simulator-only labels, including object states, geometry, robot states, and contact information. The PPO experts are used only for data generation and are discarded after dataset collection. More details please refer to Appendix~\ref{app:rl_expert}.

\subsection{Privileged Tactile Representation Learning}
\label{subsec:representation}

Stable dexterous grasping requires knowing not only whether contact occurs, but also where it occurs on the hand. Raw tactile observations $b_t \in \{0,1\}^{44}$ are flat binary taxel activations and do not explicitly encode taxel locations. We address this by (i) augmenting tactile signals with 3D taxel positions and (ii) pretraining a tactile encoder with geometric supervision.

For each tactile channel, we pair its binary activation with the 3D position of the corresponding taxel. Given the hand configuration, forward kinematics maps each taxel from its local link frame to a common hand-centric frame:
\begin{equation}
    s_\tau^i = [\mathrm{FK}_i(q_\tau; r_i), b_\tau^i] \in \mathbb{R}^{4}.
\end{equation}
We stack these kinematically grounded tactile points over all taxels and a short history window to form the tactile input $\mathcal{S}_t$. This representation preserves contact geometry while using only proprioception and binary tactile feedback, both available at deployment.

We encode $\mathcal{S}_t$ with a lightweight temporal Transformer:
\begin{equation}
    z_t = g_{\psi}(\mathcal{S}_t),
\end{equation}
where $z_t$ conditions the grasping policy. Because the input is grounded in a consistent hand-centric frame, the encoder can infer spatial contact relationships from the fixed ordering of entries in the vector.

As shown by our experiments, directly learning a policy from sparse binary contacts using imitation is difficult. In simulation, we therefore pretrain a encoder with privileged labels that are unavailable on the real robot, including object pose, object geometry, robot state, and contact annotations. Specifically, we attach a decoder to the tactile encoder and train the resulting encoder--decoder model to predict these simulator-only geometric and contact targets from tactile history. After pretraining, the decoder is discarded, and only the encoder \(g_{\psi}\) is used for policy learning and deployment. Notably, privileged information is used only as simulation-time supervision; the deployed policy still receives only proprioception and binary tactile observations. Pretraining details, architectures, and loss definitions are provided in Appendix~\ref{sec:pretrain_details}.

{
\subsection{Policy Learning and Real-World Deployment}
\label{subsec:policy}

Blind tactile grasping is inherently multimodal in action distribution: the same sparse contact history can correspond to different object poses, shapes, and feasible grasp motions. We model this ambiguity with a tactile-conditioned Diffusion Policy~\cite{chi2024diffusionpolicyvisuomotorpolicy}. At each control step $t$, the pretrained tactile encoder maps the spatially grounded tactile history $\mathcal{S}_t$ to a feature $z_t$, which is concatenated with the proprioceptive state $x_t$:
\begin{equation}
    c_t = [z_t, x_t].
\end{equation}
Conditioned on $c_t$, the policy predicts a future action chunk,
\begin{equation}
    A_t = [a_t, a_{t+1}, \ldots, a_{t+T_p-1}],
\end{equation}
and is trained on expert action chunks from the simulated demonstration dataset $\mathcal{D}$ using the standard diffusion denoising objective.

At deployment, the policy runs as a closed-loop receding-horizon controller on the physical robot. At each step, the robot updates proprioception and tactile history, computes $z_t$, samples an action chunk, and executes the first few actions before replanning. Actions command incremental arm and hand joint motions tracked by low-level joint controllers. This closed-loop execution lets the policy react to new tactile contacts, refine finger coordination, and lift objects without vision or object pose estimates. Implementation details are provided in Appendix~\ref{app:policy_details}.

}




\section{Experiments}

In this section, we evaluate whether our proposed policy enables blind grasping (i.e., grasping without visual input, object pose estimates, or geometry priors). We conduct simulation and real-world experiments to assess end-to-end grasping performance and compare the full system with ablated variants to study: 
1) whether the calibrated simulator reproduces real tactile observations, and 
2) whether tactile encoder pretraining improves policy learning.

We use a benchmark of 20 objects: 10 seen and 10 unseen. The seen objects are used only in simulation for expert demonstration generation, tactile encoder pretraining, and diffusion policy training; no real-world data is used for training. The unseen objects are held out from all training stages and used exclusively for evaluation, testing zero-shot generalization to novel objects. Additional setup details and metric definitions are provided in Appendix~\ref{sec:experiment_detail_and_Metrics}.

\begin{figure*}[t]
    \centering
    \includegraphics[width=\textwidth]{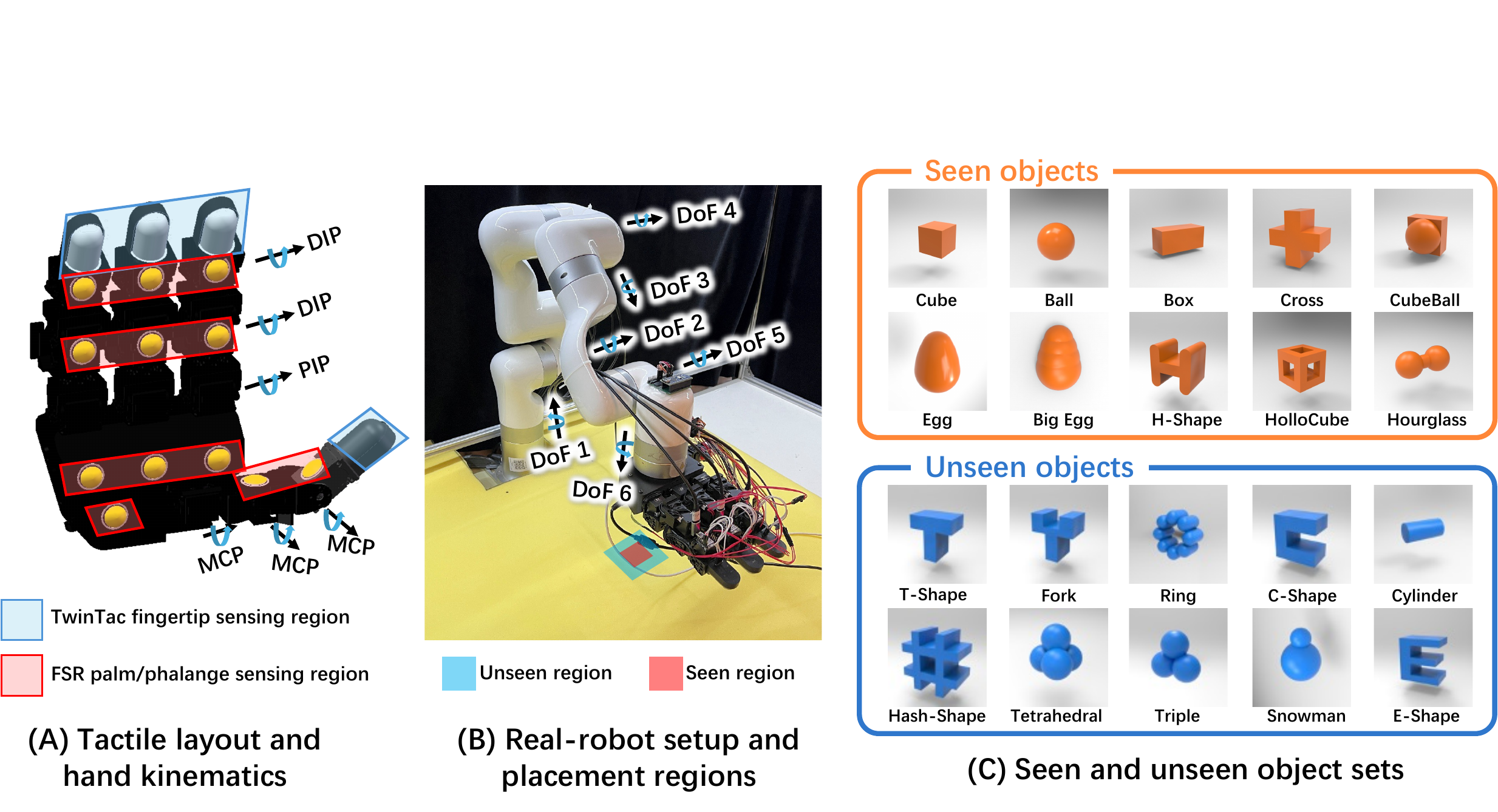}
    \caption{
Hardware setup, tactile layout, and object sets for blind dexterous grasping.
(A) Tactile sensing layout on the 16-DoF LEAP Hand. Four TwinTac sensors are mounted on the fingertips, and FSR sensors are distributed over the palm and phalanges, forming a 44-channel binary tactile observation together with proprioceptive feedback.
(B) Physical xArm6--LEAP Hand platform used for real-world deployment. The colored regions indicate training and held-out object-placement regions for evaluating spatial generalization under blind tactile control.
(C) Object sets used in the experiments. Seen objects are used for simulated demonstration generation and policy learning, while unseen objects are held out to evaluate shape generalization.
}
    \label{fig:experiment_overview}
    \vspace{-20px}
\end{figure*}

\textbf{Q1: Does the complete system enable real-world blind grasping?}
We first evaluate whether the complete system can effectively achieve blind grasping of real-world objects. The evaluation results are shown in Table~\ref{tab:real_seen_unseen}. Specifically, our method achieves a 27\% overall success rate, with successful blind grasps on both seen and unseen objects. It is also noteworthy that when objects are not centered in the hand, the policy exhibits tactile exploration behavior to align the object with the palm. Successful and failure demonstrations can be found in our supplementary video.

\begin{table*}[t]
\centering
\small
\setlength{\tabcolsep}{6pt}
\renewcommand{\arraystretch}{1.15}
\caption{
Per-object real-world blind grasping success of our full system.
All results use the tactile-calibrated simulator and privileged tactile encoder pretraining.
}
\label{tab:real_seen_unseen}
\begin{tabular}{lc lc | lc lc}
\toprule
\multicolumn{4}{c|}{\textbf{Seen Objects}}
&
\multicolumn{4}{c}{\textbf{Unseen Objects}}
\\
\cmidrule(lr){1-4}
\cmidrule(lr){5-8}
\textbf{Object} & \textbf{Success}
& \textbf{Object} & \textbf{Success}
&
\textbf{Object} & \textbf{Success}
& \textbf{Object} & \textbf{Success}
\\
\midrule
Cube & 4/5
& Ball & 0/5
& Hash-Shape & 1/5
& C-Shape & 1/5
\\
Box & 1/5
& Cross & 1/5
& E-Shape & 1/5
& T-Shape & 2/5
\\
CubeBall & 1/5
& Egg & 1/5
& Cylinder & 0/5
& Fork & 3/5
\\
Big Egg & 1/5
& H-shape & 2/5
& Ring & 1/5
& Snowman & 1/5
\\
Hollow Cube & 4/5
& Hourglass & 1/5
& Tetrahedral & 1/5
& Triple & 0/5
\\
\midrule
\multicolumn{4}{c|}{\textbf{Seen Overall: 16/50}}
&
\multicolumn{4}{c}{\textbf{Unseen Overall: 11/50}}
\\
\bottomrule
\vspace{-10px}
\end{tabular}
\end{table*}

\textbf{Q2: Does Real2Sim calibration improve tactile event fidelity?}
Reliable sim-to-real transfer requires simulated tactile events to match real sensor events in both timing and taxel location. Contacts that occur too early, too late, or at incorrect taxels can lead the policy to learn contact-conditioned actions that do not transfer to hardware.
We therefore validate the fidelity of tactile simulation before and after Real2Sim calibration. Specifically, we measure whether the calibrated simulator produces tactile observations that better match the physical sensors. We use three metrics: contact onset error, activation F1, and false positive rate (FPR). Contact onset error measures the temporal alignment of the first contact event; activation F1 measures taxel-level agreement between simulated and real contact patterns; and FPR measures erroneous simulated activations when the real sensor is inactive. Metric definitions are provided in Appendix~\ref{app:tactile_metrics}.
As shown in Table~\ref{tab:real_calibration}, the calibrated simulator achieves a low contact onset error, a higher activation F1 score, and a lower false positive rate. These results indicate that the calibrated simulator improves tactile event fidelity, supporting downstream policy learning and real-world transfer.

\textbf{Q3: Does privileged pretraining improve tactile representation learning?}
Sparse binary tactile events provide intermittent, weakly structured information about hand--object interactions. Without the additional priors provided by pretraining, the policy must learn both tactile representations and grasping behavior from the downstream imitation loss alone, making representation learning extremely difficult and inefficient.

Keeping the tactile input, policy architecture, and demonstration data fixed, we compare policies trained with and without privileged pretraining. For a controlled evaluation, we conduct these experiments in simulation to isolate variability from the real world. As shown in Table~\ref{tab:sim_pretrain_compact}, pretraining improves success on both seen and unseen objects, indicating that simulator-only geometric supervision helps the encoder learn more structured representations of sparse contact events before policy learning. This trend is consistent with the hardware results, where removing privileged pretraining reduces real-world grasping performance. Together, these results show that privileged pretraining improves tactile representation learning and leads to more robust blind grasping after transfer to physical robot hand.

\paragraph{Summary and Additional Ablations.}
Taken together, the experiments demonstrate successful grasping and support the design choices used in the deployed system. Real2Sim calibration improves the simulator's tactile event fidelity, while the pretraining ablations show that privileged tactile encoder pretraining improves policy learning from sparse tactile observations. Additional ablations, including real-world pretraining results, policy architecture, tactile signal representation, and kinematically grounded tactile inputs, are provided in Appendix~\ref{app:ablation_experiments}.

\begin{table*}[t]
\centering
\small
\setlength{\tabcolsep}{4pt}
\renewcommand{\arraystretch}{1.12}

\begin{minipage}[t]{0.48\textwidth}
\centering
\vspace{0pt}
\captionof{table}{Effect of Real2Sim tactile calibration.}
\label{tab:real_calibration}
\begin{tabular}{lccc}
\toprule
\textbf{Method}
& $\Delta t_{\mathrm{on}}$ (ms)
& F1 (\%) 
& FPR (\%)  \\
\midrule
Nominal Sim.
& 426 & 64.2 & 4.2 \\
\rowcolor{blue!5}\textbf{Calibrated Sim.}
& \textbf{96} & \textbf{70.5} & \textbf{2.2} \\
\bottomrule
\end{tabular}
\end{minipage}
\hfill
\begin{minipage}[t]{0.50\textwidth}
\centering
\vspace{0pt}
\captionof{table}{Effect of tactile encoder pretraining.}
\label{tab:sim_pretrain_compact}
\begin{tabular}{lccc}
\toprule
\textbf{Method}
& \textbf{Seen }
& \textbf{Unseen }
& \textbf{Overall} \\
\midrule
w/o Pretrain
& 36.2\% & 20.0\% & 28.1\% \\
\rowcolor{blue!5}\textbf{w/ Pretrain}
& \textbf{60.4\%} & \textbf{43.2\%} & \textbf{51.8\%} \\
\bottomrule
\end{tabular}
\end{minipage}

\vspace{-10pt}
\end{table*}


\section{Limitations}

Our system has several limitations. First, the overall success rate is 27\%. While this remains modest, the evaluation is challenging because it uses a large and diverse object set. To better understand these challenges, we provide a comprehensive failure mode analysis in Appendix~\ref{sec:appendix_failure_mode}.

From our observations, many failures occur because the policy does not establish stable grasp points within the fixed execution horizon. Although the hand may initially contact the object, subsequent finger motions do not always reposition the contacts into stable grasp configurations, leading to empty grasps or object slip/drop during lifting. These failures are observed in both simulation and the real system, suggesting that they arise from limitations in the policy, the hardware, or both. Moreover, the current hardware provides incomplete tactile coverage. During interaction, contacts frequently occur in unsensed regions of the hand, causing the policy to execute actions based on partial or ambiguous contact observations. This limitation likely contributes to unstable grasp formation and reduced robustness.

These observations suggest several directions for future work. One promising direction is improving tactile sensing quality and coverage. Integrating full-hand tactile skins~\cite{Zhao_2025} could provide denser spatial feedback and reduce failures caused by contacts in currently unsensed regions. Incorporating explicit shear or slip sensing may further enable earlier detection of incipient grasp failure and reduce spatial aliasing in tactile observations.

Another direction is improving control and policy design. Tactile-only grasping may benefit from a two-timescale control architecture consisting of a slower high-level policy for contact search and grasp formation, combined with a faster tactile reflex or servo controller for grip-force regulation and slip recovery. Such a reactive layer could be implemented either as a learned module or as an adaptive low-level controller, and may help bridge the residual contact-dynamics gap between simulation and the physical world. Future work could also improve grasp stability through richer tactile representations, stronger contact or object priors, recovery and regrasp behaviors, and more diverse simulated demonstrations.


\section{Conclusions}
This paper presents a complete pipeline for learning and deploying a blind grasping policy on a robotic dexterous hand. Our pipeline combines a tactile-enabled dexterous hand platform, a high-fidelity tactile simulation environment calibrated via Real2Sim, and an RL expert policy with a distillation framework for Sim2Real deployment. Through experiments and ablations, we find that successful blind grasping depends critically on aligning contact-event observations between simulation and the real system, as well as incorporating a pretrained tactile encoder into the policy. By jointly leveraging these components, our system achieves a 27\% grasp success rate on a large and diverse real-world object set. Future work will focus on improving grasp robustness through richer tactile representations, recovery and regrasp behaviors, and denser full-hand tactile sensing.



\clearpage
\acknowledgments{This work was supported by the National Natural Science Foundation of China (Grant No. 52305007), the Natural Science Foundation of Shanghai (Grant No. 25ZR1402370), the Artificial Intelligence Project of the State Key Laboratory of General Artificial Intelligence, BIGAI, Peking University, Beijing, China (Project No. SKLAGI20250P19), the State Key Laboratory of Mechanical System and Vibration (Grant No. MSV202519) and the MoE Key Laboratory of Intelligent Perception and Human-Machine Collaboration (KLIP-HuMaCo).}


\bibliography{example}  

@misc{zhao2023learningfinegrainedbimanualmanipulation,
      title={Learning Fine-Grained Bimanual Manipulation with Low-Cost Hardware}, 
      author={Tony Z. Zhao and Vikash Kumar and Sergey Levine and Chelsea Finn},
      year={2023},
      eprint={2304.13705},
      archivePrefix={arXiv},
      primaryClass={cs.RO},
      url={https://arxiv.org/abs/2304.13705}, 
}

@article{2018More,
  title={More Than a Feeling: Learning to Grasp and Regrasp using Vision and Touch},
  author={ Roberto, Calandra  and  Andrew, Owens  and  Dinesh, Jayaraman  and  Justin, Lin  and  Wenzhen, Yuan  and  Jitendra, Malik  and  Adelson, Edward H.  and  Sergey, Levine },
  journal={IEEE Robotics \& Automation Letters},
  volume={3},
  number={4},
  pages={3300-3307},
  year={2018},
}

@misc{yang2024anyrotategravityinvariantinhandobject,
      title={AnyRotate: Gravity-Invariant In-Hand Object Rotation with Sim-to-Real Touch}, 
      author={Max Yang and Chenghua Lu and Alex Church and Yijiong Lin and Chris Ford and Haoran Li and Efi Psomopoulou and David A. W. Barton and Nathan F. Lepora},
      year={2024},
      eprint={2405.07391},
      archivePrefix={arXiv},
      primaryClass={cs.RO},
      url={https://arxiv.org/abs/2405.07391}, 
}

@misc{loshchilov2019decoupledweightdecayregularization,
      title={Decoupled Weight Decay Regularization}, 
      author={Ilya Loshchilov and Frank Hutter},
      year={2019},
      eprint={1711.05101},
      archivePrefix={arXiv},
      primaryClass={cs.LG},
      url={https://arxiv.org/abs/1711.05101}, 
}

@misc{zakka2025mujocoplayground,
      title={MuJoCo Playground}, 
      author={Kevin Zakka and Baruch Tabanpour and Qiayuan Liao and Mustafa Haiderbhai and Samuel Holt and Jing Yuan Luo and Arthur Allshire and Erik Frey and Koushil Sreenath and Lueder A. Kahrs and Carmelo Sferrazza and Yuval Tassa and Pieter Abbeel},
      year={2025},
      eprint={2502.08844},
      archivePrefix={arXiv},
      primaryClass={cs.RO},
      url={https://arxiv.org/abs/2502.08844}, 
}

@article{Liu_2022,
   title={Synthesizing Diverse and Physically Stable Grasps With Arbitrary Hand Structures Using Differentiable Force Closure Estimator},
   volume={7},
   ISSN={2377-3774},
   url={http://dx.doi.org/10.1109/LRA.2021.3129138},
   DOI={10.1109/lra.2021.3129138},
   number={1},
   journal={IEEE Robotics and Automation Letters},
   publisher={Institute of Electrical and Electronics Engineers (IEEE)},
   author={Liu, Tengyu and Liu, Zeyu and Jiao, Ziyuan and Zhu, Yixin and Zhu, Song-Chun},
   year={2022},
   month=Jan, pages={470–477} }

@article{Chen2024GeneralPurposeSP,
  title={General-Purpose Sim2Real Protocol for Learning Contact-Rich Manipulation With Marker-Based Visuotactile Sensors},
  author={Weihang Chen and Jing Xu and Fanbo Xiang and Xiao Yuan and Hao Su and Rui Chen},
  journal={IEEE Transactions on Robotics},
  year={2024},
  volume={40},
  pages={1509-1526},
  url={https://api.semanticscholar.org/CorpusID:267005473}
}

@article{2018The,
  title={The TacTip Family: Soft Optical Tactile Sensors with 3D-Printed Biomimetic Morphologies},
  author={ Ward-Cherrier, Benjamin  and  Pestell, Nicholas  and  Cramphorn, Luke  and  Winstone, Benjamin  and  Giannaccini, Maria Elena  and  Rossiter, Jonathan  and  Lepora, Nathan F. },
  journal={Soft Robotics},
  pages={216-227},
  year={2018},
}

@inproceedings{Church2021TactileSP,
  title={Tactile Sim-to-Real Policy Transfer via Real-to-Sim Image Translation},
  author={Alex Church and John Lloyd and Raia Hadsell and Nathan F. Lepora},
  booktitle={Conference on Robot Learning},
  year={2021},
  url={https://api.semanticscholar.org/CorpusID:240354208}
}

@article{Lach2023TowardsTT,
  title={Towards Transferring Tactile-based Continuous Force Control Policies from Simulation to Robot},
  author={Luca Lach and Robert Haschke and Davide Tateo and Jan Peters and Helge J. Ritter and J{\'u}lia Borr{\`a}s Sol and Carme Torras},
  journal={ArXiv},
  year={2023},
  volume={abs/2311.07245},
  url={https://api.semanticscholar.org/CorpusID:265149679}
}

@article{touch-dexterity,
      title          = {Rotating without Seeing: Towards In-hand Dexterity through Touch },
      author         = {Yin, Zhao-Heng and Huang, Binghao and Qin, Yuzhe and Chen, Qifeng and Wang, Xiaolong},
      journal        = {Robotics: Science and Systems},
      year           = {2023},
    }

@inproceedings{conf/icra/DangWA11,
  author = {Dang, Hao and Weisz, Jonathan and Allen, Peter K.},
  booktitle = {ICRA},
  ee = {http://dx.doi.org/10.1109/ICRA.2011.5979679},
  pages = {5917-5922},
  publisher = {IEEE},
  title = {Blind grasping: Stable robotic grasping using tactile feedback and hand kinematics.},
  url = {http://dblp.uni-trier.de/db/conf/icra/icra2011.html#DangWA11},
  year = 2011
}

@misc{huang20253dvitaclearningfinegrainedmanipulation,
      title={3D-ViTac: Learning Fine-Grained Manipulation with Visuo-Tactile Sensing}, 
      author={Binghao Huang and Yixuan Wang and Xinyi Yang and Yiyue Luo and Yunzhu Li},
      year={2025},
      eprint={2410.24091},
      archivePrefix={arXiv},
      primaryClass={cs.RO},
      url={https://arxiv.org/abs/2410.24091}, 
}

@misc{fei2025trograspefficientgraph,
      title={T(R,O) Grasp: Efficient Graph Diffusion of Robot-Object Spatial Transformation for Cross-Embodiment Dexterous Grasping}, 
      author={Xin Fei and Zhixuan Xu and Huaicong Fang and Tianrui Zhang and Lin Shao},
      year={2025},
      eprint={2510.12724},
      archivePrefix={arXiv},
      primaryClass={cs.RO},
      url={https://arxiv.org/abs/2510.12724}, 
}

@ARTICLE{10246361,
  author={Li, Hongyu and Dikhale, Snehal and Iba, Soshi and Jamali, Nawid},
  journal={IEEE Robotics and Automation Letters}, 
  title={ViHOPE: Visuotactile In-Hand Object 6D Pose Estimation With Shape Completion}, 
  year={2023},
  volume={8},
  number={11},
  pages={6963-6970},
  doi={10.1109/LRA.2023.3313941}}

@misc{lee2025vitascopevisuotactileimplicitrepresentation,
      title={ViTaSCOPE: Visuo-tactile Implicit Representation for In-hand Pose and Extrinsic Contact Estimation}, 
      author={Jayjun Lee and Nima Fazeli},
      year={2025},
      eprint={2506.12239},
      archivePrefix={arXiv},
      primaryClass={cs.RO},
      url={https://arxiv.org/abs/2506.12239}, 
}

@article{lin2024twisting,
          author={Lin, Toru and Yin, Zhao-Heng and Qi, Haozhi and Abbeel, Pieter and Malik, Jitendra},
          title={Twisting Lids Off with Two Hands},
          journal={arXiv:2403.02338},
          year={2024}
        }

@inproceedings{
    xu2022efficient,
    title={Efficient Tactile Simulation with Differentiability for Robotic Manipulation},
    author={Jie Xu and Sangwoon Kim and Tao Chen and Alberto Rodriguez Garcia and Pulkit Agrawal and Wojciech Matusik and Shinjiro Sueda},
    booktitle={6th Annual Conference on Robot Learning},
    year={2022},
    url={https://openreview.net/forum?id=6BIffCl6gsM}
    }

@misc{si2021taximexamplebasedsimulationmodel,
      title={Taxim: An Example-based Simulation Model for GelSight Tactile Sensors}, 
      author={Zilin Si and Wenzhen Yuan},
      year={2021},
      eprint={2109.04027},
      archivePrefix={arXiv},
      primaryClass={cs.RO},
      url={https://arxiv.org/abs/2109.04027}, 
}

@inproceedings{
li2025taccel,
title={Taccel: Scaling Up Vision-based Tactile Robotics via High-performance {GPU} Simulation},
author={Yuyang Li and Wenxin Du and Chang Yu and Puhao Li and Zihang Zhao and Tengyu Liu and Chenfanfu Jiang and Yixin Zhu and Siyuan Huang},
booktitle={The Thirty-ninth Annual Conference on Neural Information Processing Systems},
year={2025},
url={https://openreview.net/forum?id=PtGMadeONU}
}

@misc{yin2023rotatingseeinginhanddexterity,
      title={Rotating without Seeing: Towards In-hand Dexterity through Touch}, 
      author={Zhao-Heng Yin and Binghao Huang and Yuzhe Qin and Qifeng Chen and Xiaolong Wang},
      year={2023},
      eprint={2303.10880},
      archivePrefix={arXiv},
      primaryClass={cs.RO},
      url={https://arxiv.org/abs/2303.10880}, 
}

@inproceedings{
church2021tactile,
title={Tactile Sim-to-Real Policy Transfer via Real-to-Sim Image Translation},
author={Alex Church and John Lloyd and raia hadsell and Nathan F. Lepora},
booktitle={5th Annual Conference on Robot Learning },
year={2021},
url={https://openreview.net/forum?id=2NcPgLa7yqD}
}

@misc{huang2026spatiallyanchoredtactileawareness,
      title={Spatially anchored Tactile Awareness for Robust Dexterous Manipulation}, 
      author={Jialei Huang and Yang Ye and Yuanqing Gong and Xuezhou Zhu and Yang Gao and Kaifeng Zhang},
      year={2026},
      eprint={2510.14647},
      archivePrefix={arXiv},
      primaryClass={cs.RO},
      url={https://arxiv.org/abs/2510.14647}, 
}

@article{chen2025dexonomy,
        title={Dexonomy: Synthesizing All Dexterous Grasp Types in a Grasp Taxonomy},
        author={Chen, Jiayi and Ke, Yubin and Peng, Lin and Wang, He},
        journal={Robotics: Science and Systems},
        year={2025}
      }

@misc{chen2025bodexscalableefficientrobotic,
      title={BODex: Scalable and Efficient Robotic Dexterous Grasp Synthesis Using Bilevel Optimization}, 
      author={Jiayi Chen and Yubin Ke and He Wang},
      year={2025},
      eprint={2412.16490},
      archivePrefix={arXiv},
      primaryClass={cs.RO},
      url={https://arxiv.org/abs/2412.16490}, 
}

@misc{bauza2020tactileobjectposeestimation,
      title={Tactile Object Pose Estimation from the First Touch with Geometric Contact Rendering}, 
      author={Maria Bauza and Eric Valls and Bryan Lim and Theo Sechopoulos and Alberto Rodriguez},
      year={2020},
      eprint={2012.05205},
      archivePrefix={arXiv},
      primaryClass={cs.RO},
      url={https://arxiv.org/abs/2012.05205}, 
}

@misc{narang2021simtorealrobotictactilesensing,
      title={Sim-to-Real for Robotic Tactile Sensing via Physics-Based Simulation and Learned Latent Projections}, 
      author={Yashraj Narang and Balakumar Sundaralingam and Miles Macklin and Arsalan Mousavian and Dieter Fox},
      year={2021},
      eprint={2103.16747},
      archivePrefix={arXiv},
      primaryClass={cs.RO},
      url={https://arxiv.org/abs/2103.16747}, 
}

@article{10.1016/j.robot.2015.07.015,
author = {Kappassov, Zhanat and Corrales, Juan-Antonio and Perdereau, V\'{e}ronique},
title = {Tactile sensing in dexterous robot hands - Review},
year = {2015},
issue_date = {December 2015},
publisher = {North-Holland Publishing Co.},
address = {NLD},
volume = {74},
number = {PA},
issn = {0921-8890},
url = {https://doi.org/10.1016/j.robot.2015.07.015},
doi = {10.1016/j.robot.2015.07.015},
journal = {Robot. Auton. Syst.},
month = dec,
pages = {195–220},
numpages = {26}
}

@misc{garciagarcia2019tactilegcngraphconvolutionalnetwork,
      title={TactileGCN: A Graph Convolutional Network for Predicting Grasp Stability with Tactile Sensors}, 
      author={Alberto Garcia-Garcia and Brayan Stiven Zapata-Impata and Sergio Orts-Escolano and Pablo Gil and Jose Garcia-Rodriguez},
      year={2019},
      eprint={1901.06181},
      archivePrefix={arXiv},
      primaryClass={cs.LG},
      url={https://arxiv.org/abs/1901.06181}, 
}

@misc{ding2021simtorealtransferroboticmanipulation,
      title={Sim-to-Real Transfer for Robotic Manipulation with Tactile Sensory}, 
      author={Zihan Ding and Ya-Yen Tsai and Wang Wei Lee and Bidan Huang},
      year={2021},
      eprint={2103.00410},
      archivePrefix={arXiv},
      primaryClass={cs.RO},
      url={https://arxiv.org/abs/2103.00410}, 
}

@INPROCEEDINGS{8246881,
  author={Yamaguchi, Akihiko and Atkeson, Christopher G.},
  booktitle={2017 IEEE-RAS 17th International Conference on Humanoid Robotics (Humanoids)}, 
  title={Implementing tactile behaviors using FingerVision}, 
  year={2017},
  volume={},
  number={},
  pages={241-248},
  doi={10.1109/HUMANOIDS.2017.8246881}}

@INPROCEEDINGS{7139016,
  author={Yuan, Wenzhen and Li, Rui and Srinivasan, Mandayam A. and Adelson, Edward H.},
  booktitle={2015 IEEE International Conference on Robotics and Automation (ICRA)}, 
  title={Measurement of shear and slip with a GelSight tactile sensor}, 
  year={2015},
  volume={},
  number={},
  pages={304-311},
  doi={10.1109/ICRA.2015.7139016}}

@misc{yuan2024robotsynesthesiainhandmanipulation,
      title={Robot Synesthesia: In-Hand Manipulation with Visuotactile Sensing}, 
      author={Ying Yuan and Haichuan Che and Yuzhe Qin and Binghao Huang and Zhao-Heng Yin and Kang-Won Lee and Yi Wu and Soo-Chul Lim and Xiaolong Wang},
      year={2024},
      eprint={2312.01853},
      archivePrefix={arXiv},
      primaryClass={cs.RO},
      url={https://arxiv.org/abs/2312.01853}, 
}

@misc{huang2026gelslamrealtimehighfidelityrobust,
      title={GelSLAM: A Real-time, High-Fidelity, and Robust 3D Tactile SLAM System}, 
      author={Hung-Jui Huang and Mohammad Amin Mirzaee and Michael Kaess and Wenzhen Yuan},
      year={2026},
      eprint={2508.15990},
      archivePrefix={arXiv},
      primaryClass={cs.RO},
      url={https://arxiv.org/abs/2508.15990}, 
}

@INPROCEEDINGS{509205,
  author={Holweg, E.G.M. and Hoeve, H. and Jongkind, W. and Marconi, L. and Melchiorri, C. and Bonivento, C.},
  booktitle={Proceedings of IEEE International Conference on Robotics and Automation}, 
  title={Slip detection by tactile sensors: algorithms and experimental results}, 
  year={1996},
  volume={4},
  number={},
  pages={3234-3239 vol.4},
  doi={10.1109/ROBOT.1996.509205}}

@misc{hu2024learningdetectsliptactile,
      title={Learning to Detect Slip through Tactile Estimation of the Contact Force Field and its Entropy}, 
      author={Xiaohai Hu and Aparajit Venkatesh and Yusen Wan and Guiliang Zheng and Neel Jawale and Navneet Kaur and Xu Chen and Paul Birkmeyer},
      year={2024},
      eprint={2303.00935},
      archivePrefix={arXiv},
      primaryClass={cs.RO},
      url={https://arxiv.org/abs/2303.00935}, 
}

@article{Zhao_2025,
   title={Embedding high-resolution touch across robotic hands enables adaptive human-like grasping},
   volume={7},
   ISSN={2522-5839},
   url={http://dx.doi.org/10.1038/s42256-025-01053-3},
   DOI={10.1038/s42256-025-01053-3},
   number={6},
   journal={Nature Machine Intelligence},
   publisher={Springer Science and Business Media LLC},
   author={Zhao, Zihang and Li, Wanlin and Li, Yuyang and Liu, Tengyu and Li, Boren and Wang, Meng and Du, Kai and Liu, Hangxin and Zhu, Yixin and Wang, Qining and Althoefer, Kaspar and Zhu, Song-Chun},
   year={2025},
   month=June, pages={889–900} }

@article{Lee_2024,
   title={DexTouch: Learning to Seek and Manipulate Objects With Tactile Dexterity},
   volume={9},
   ISSN={2377-3774},
   url={http://dx.doi.org/10.1109/LRA.2024.3478571},
   DOI={10.1109/lra.2024.3478571},
   number={12},
   journal={IEEE Robotics and Automation Letters},
   publisher={Institute of Electrical and Electronics Engineers (IEEE)},
   author={Lee, Kang-Won and Qin, Yuzhe and Wang, Xiaolong and Lim, Soo-Chul},
   year={2024},
   month=dec, pages={10772–10779} }

@misc{shaw2023leaphandlowcostefficient,
      title={LEAP Hand: Low-Cost, Efficient, and Anthropomorphic Hand for Robot Learning}, 
      author={Kenneth Shaw and Ananye Agarwal and Deepak Pathak},
      year={2023},
      eprint={2309.06440},
      archivePrefix={arXiv},
      primaryClass={cs.RO},
      url={https://arxiv.org/abs/2309.06440}, 
}

@misc{chi2024diffusionpolicyvisuomotorpolicy,
      title={Diffusion Policy: Visuomotor Policy Learning via Action Diffusion}, 
      author={Cheng Chi and Zhenjia Xu and Siyuan Feng and Eric Cousineau and Yilun Du and Benjamin Burchfiel and Russ Tedrake and Shuran Song},
      year={2024},
      eprint={2303.04137},
      archivePrefix={arXiv},
      primaryClass={cs.RO},
      url={https://arxiv.org/abs/2303.04137}, 
}

@INPROCEEDINGS{Xiyan2025Twintac,
  author={Huang, Xiyan and Xu, Zhe and Xiao, Chenxi},
  booktitle={2025 IEEE/RSJ International Conference on Intelligent Robots and Systems (IROS)}, 
  title={TwinTac: A Wide-Range, Highly Sensitive Tactile Sensor with Real-To-Sim Digital Twin Sensor Model}, 
  year={2025},
  volume={},
  number={},
  pages={12286-12292},
  doi={10.1109/IROS60139.2025.11247002}}

@article{vinogradova2003interaction,
  title={Interaction of elastic bodies via surface forces: 2. Exponential decay},
  author={Vinogradova, Olga I and Feuillebois, Fran{\c{c}}ois},
  journal={Journal of colloid and interface science},
  volume={268},
  number={2},
  pages={464--475},
  year={2003},
  publisher={Elsevier}
}

@misc{KaolinLibrary,
      author = {Fuji Tsang, Clement and Shugrina, Maria and Lafleche, Jean Francois and Takikawa, Towaki and Wang, Jiehan and Loop, Charles et al.},
      title = {Kaolin: A Pytorch Library for Accelerating 3D Deep Learning Research},
      year = {2022},
      howpublished={\url{https://github.com/NVIDIAGameWorks/kaolin}}
}

@article{lin2025pp,
  title={Pp-tac: Paper picking using tactile feedback in dexterous robotic hands},
  author={Lin, Pei and Huang, Yuzhe and Li, Wanlin and Ma, Jianpeng and Xiao, Chenxi and Jiao, Ziyuan},
  journal={arXiv preprint arXiv:2504.16649},
  year={2025}
}

\clearpage
\appendix
\section*{Appendix Overview}
\label{app:overview}

This appendix provides additional details for hardware design, evaluation protocol, failure analysis, ablations, expert data generation, tactile encoder pretraining, and policy deployment.

\vspace{0.5em}
\noindent
\textbf{\hyperref[app:hardware]{A. Hardware and Tactile Sensing System}}  
describes the tactile hardware, including the customized TwinTac fingertip sensors and FSR patches on the palm and links. 
It also details the Real2Sim tactile calibration procedure and the construction of the binary tactile observation used by the policy.

\vspace{0.35em}
\noindent
\textbf{\hyperref[sec:experiment_detail_and_Metrics]{B. Evaluation Protocol and Metrics}}  
details the grasp evaluation setup, success criterion, and tactile event fidelity metrics used to compare real and simulated tactile streams.

\vspace{0.35em}
\noindent
\textbf{\hyperref[app:ablation_experiments]{C. Additional Ablation Studies}}  
reports additional simulation and real-world ablations on privileged tactile encoder pretraining, tactile signal representation, kinematically grounded tactile inputs, and policy architecture.

\vspace{0.35em}
\noindent
\textbf{\hyperref[sec:appendix_failure_mode]{D. Detailed Real-World Failure Mode Analysis}}  
analyzes the dominant failure modes observed in real-world trials, including empty grasps, slips during lifting, and emergency stops.

\vspace{0.35em}
\noindent
\textbf{\hyperref[app:rl_expert]{E. RL Expert Training Details}}  
describes the PPO experts used for demonstration collection, including actor and critic observations, task and domain randomization, and PPO hyperparameters.

\vspace{0.35em}
\noindent
\textbf{\hyperref[sec:pretrain_details]{F. Privileged Tactile Encoder Pretraining Details}}  
provides details on the tactile input representation, Transformer encoder, privileged decoder targets, pretraining losses, and training hyperparameters.

\vspace{0.35em}
\noindent
\textbf{\hyperref[app:policy_details]{G. Diffusion Policy Training and Deployment Details}}  
describes the diffusion policy training objective, receding-horizon execution, and action parameterization used for real-world deployment.

\vspace{1em}

\section{Hardware and Tactile Sensing System}
\label{app:hardware}

To provide tactile perception for dexterous manipulation, we augment the Leap Hand with a hybrid tactile sensing system consisting of four customized TwinTac fingertip sensors~\cite{Xiyan2025Twintac} and 12 distributed force-sensitive resistor (FSR) patches mounted on the finger links and palm. Each TwinTac sensor contains eight pressure-sensitive taxels, providing 32 fingertip tactile channels in total. We use TwinTac sensors and FSRs because they are low-cost, customizable, and vision-free (i.e., they do not require the visual processing pipeline used by vision-based tactile sensors). These properties make them easy to integrate with the Leap Hand geometry. The sensor integration details are described below.

\begin{figure}[!ht]
    \centering
    \includegraphics[width=1.0\linewidth]{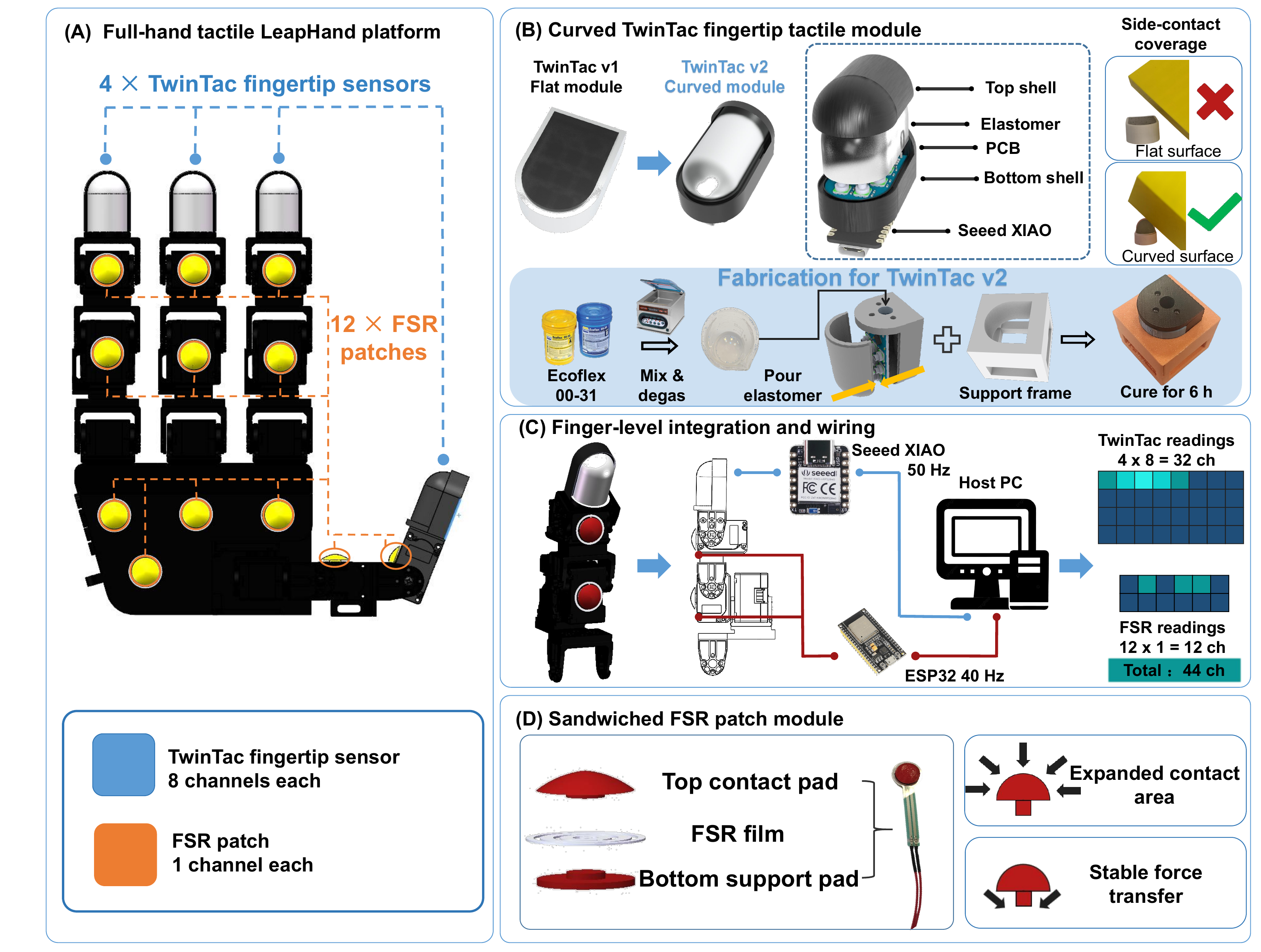}
    \caption{\textbf{Full-hand tactile LeapHand platform.}
The platform integrates four curved TwinTac fingertip sensors and twelve distributed FSR patches, forming a 44-channel tactile sensing system. 
The curved TwinTac module provides high-resolution fingertip contact measurements through an 8-taxel pressure array, while the FSR patches extend tactile coverage to non-fingertip regions including finger links and the palm. 
The tactile signals are routed through the acquisition electronics and streamed to the host computer for policy execution.
}
    \label{fig:TactileLeap}
\end{figure}

\subsection{Customized Fingertip TwinTac Sensors}

We customize the TwinTac~\cite{Xiyan2025Twintac} fingertip sensors in two main ways. First, we recast the gel elastomer into a curved sensing surface, which increases the effective contact area compared with the original flat TwinTac design. This is important for multi-finger manipulation, where contacts frequently occur on the side regions of the fingertip rather than only on the frontal surface. Second, we integrate an Seeed XIAO microcontroller on the rear side of each sensor, enabling direct USB communication via VCP UART instead of the original I2C interface. This simplifies wiring and reduces cable interference with finger motion.

To gel-cast the curved elastomer, we use three 3D-printed PLA parts: a bottom sensor shell, a removable top mold, and a vertical casting fixture. The TwinTac PCB is fixed inside the bottom shell, and the top mold is aligned with it to define the curved fingertip geometry. The assembly is then placed vertically in the casting fixture, and mixed EcoFlex 00-31 is poured from the top opening and cured in place. After curing, the sensor is removed from the fixture, and the top mold is detached through the reserved edge, leaving a curved elastomeric sensing surface integrated with the bottom shell. Rear screw holes allow direct mounting onto the Leap Hand fingertip transmission element while preserving the tactile sensing surface.

\subsection{Palm and Link FSR Sensors}

For non-fingertip contact sensing, we use 12 low-cost force-sensitive resistor (FSR) patches. Each FSR measures local normal pressure through a voltage-divider circuit. To improve mechanical coupling on the curved Leap Hand servo housings, each patch is sandwiched between two 3D-printed layers: a bottom layer for stable mounting and a top cap for force transmission, a larger effective contact area, and improved contact conformity (Fig.~\ref{fig:TactileLeap}). This compact assembly fits on the finger links and palm.
Finally, the 12 FSR voltage signals are digitized by an STM32 microcontroller, streamed to the host via serial at approximately 40\,Hz, and thresholded into binary contact indicators for each sensing location.

\begin{figure}[!ht]
    \centering
    \includegraphics[width=1\linewidth]{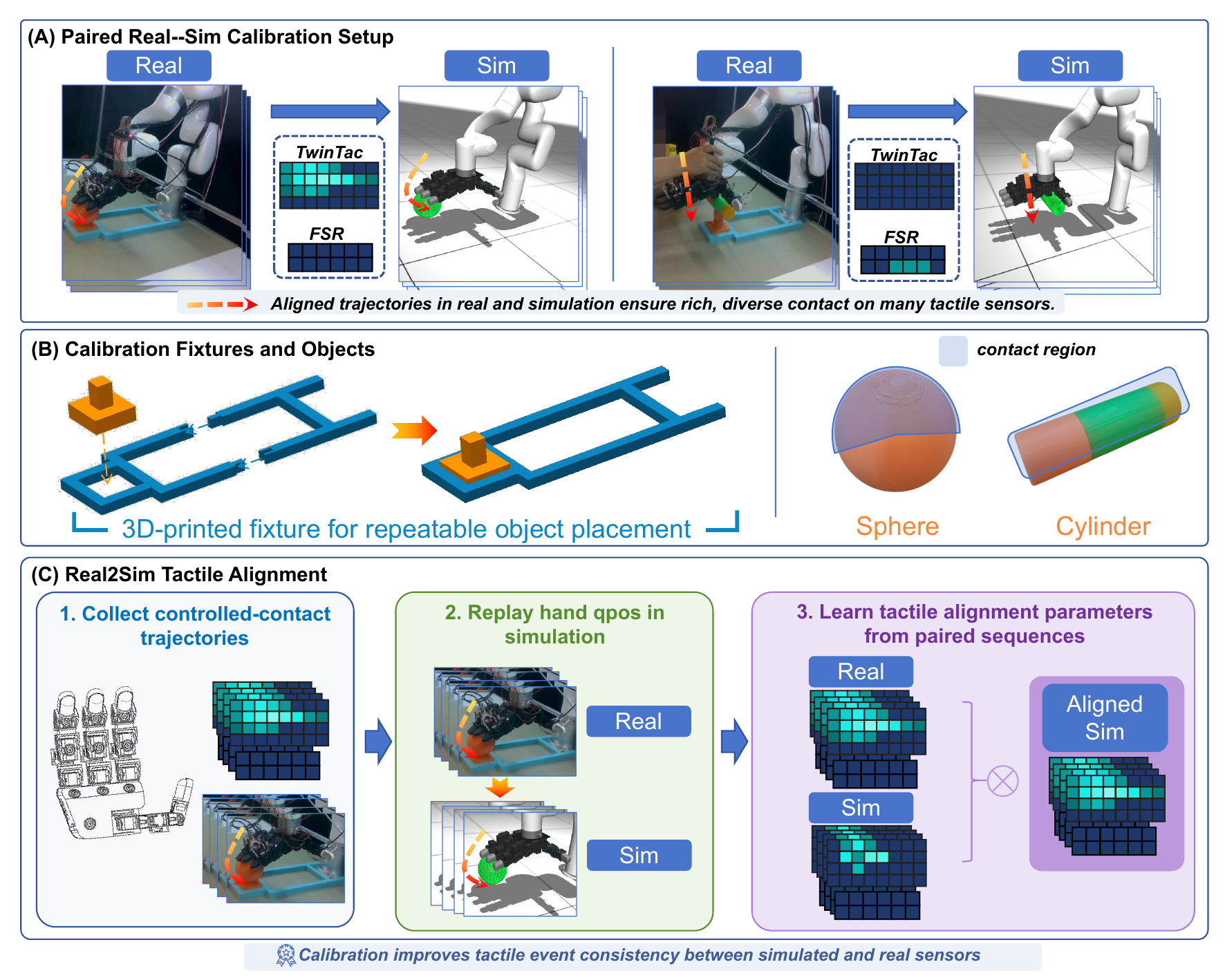}
    \caption{
    \textbf{Real2Sim tactile alignment system.}
(A) Paired calibration trajectories are collected in the real setup and replayed in simulation under matched object poses. During real-world collection, the hand interacts with fixed calibration objects while raw tactile readings and hand joint positions (\textit{qpos}) are recorded; the recorded \textit{qpos} are then replayed in simulation to obtain temporally aligned real--sim tactile sequences.
(B) 3D-printed fixtures are used to place the calibration objects at repeatable poses consistent with the simulator. We use two simple calibration geometries: a sphere with radius $5\,\mathrm{cm}$ and a cylinder with radius $2.5\,\mathrm{cm}$ and height $10\,\mathrm{cm}$.
(C) The paired sequences are used to learn the tactile alignment network $g_\phi$, which maps raw real tactile readings to simulation-compatible contact features. The calibration motions are task-agnostic controlled-contact trajectories rather than grasping demonstrations or policy rollouts. In total, we collect nine calibration trajectories across the two objects at approximately $10\,\mathrm{Hz}$, using arm lead-through and UI-based hand joint control to induce contacts across different tactile regions.
    }
    \label{fig:calibrationSys}
\end{figure}

\subsection{Tactile Signal Calibration and Observation Construction}

Our calibration in main paper is performed as follows. We first subtract the initial reference values from all raw tactile readings. During policy execution, the corrected signals are binarized and concatenated in a fixed order: the first 32 entries correspond to TwinTac measurements, and the remaining 12 entries correspond to FSR measurements. The resulting tactile vector is included in the policy observation.

For calibration, we use two objects with simple geometries: a sphere of radius $5\,\mathrm{cm}$ and a cylinder of radius $2.5\,\mathrm{cm}$ and height $10\,\mathrm{cm}$. We collect nine calibration trajectories with varying durations. The tactile and joint-state data are recorded at approximately $10\,\mathrm{Hz}$. In total, the sphere trajectories contain $325.5\,\mathrm{s}$ of calibration data, corresponding to $3{,}255$ frames, while the cylinder trajectories contain $241.9\,\mathrm{s}$ of calibration data, corresponding to $2{,}419$ frames. This results in $5{,}674$ calibration frames across the two objects. We then replay the recorded hand joint positions in simulation to obtain temporally aligned real--sim tactile sequences for parameter fitting and sensing-level evaluation.

The calibration trajectories are collected semi-manually to ensure diverse tactile excitation. The robot arm is moved in teleoperation mode, allowing the operator to guide the hand around the fixed calibration object. Meanwhile, the tactile collection process follows a two-stage control sequence. The robot is first manually guided to a starting pose via a standard user interface. Once positioned, it executes a predetermined trajectory designed to bring specific tactile regions into controlled contact with the object. This procedure provides practical coverage of different tactile regions while keeping the calibration data independent of the downstream grasping policy.

\begin{figure}[t]
    \centering
    \includegraphics[width=1\linewidth]{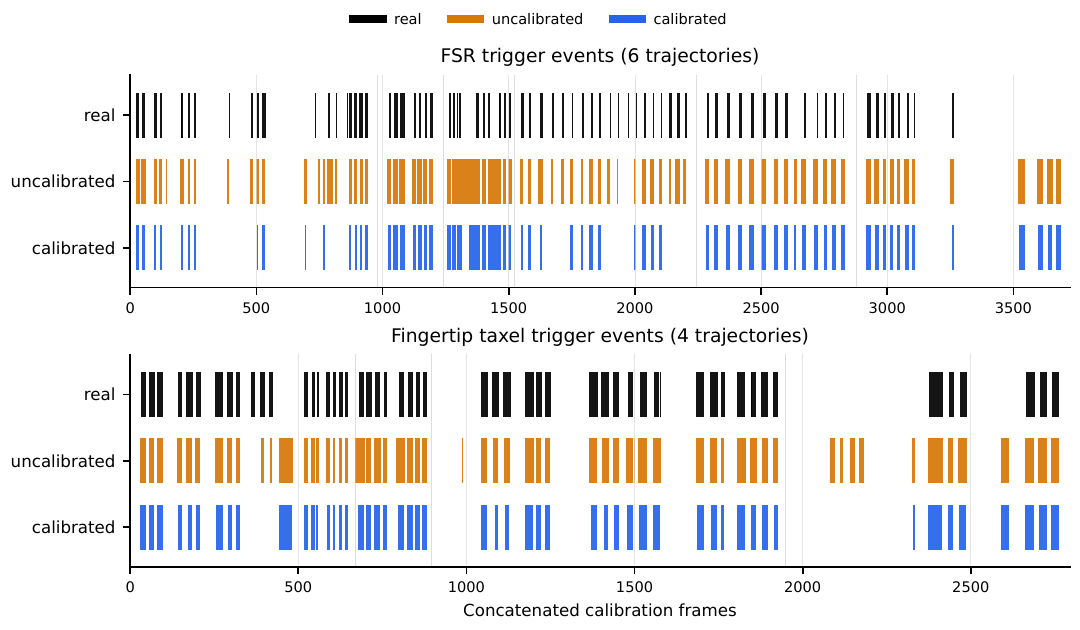}
    \caption{
    \textbf{Tactile activation alignment before and after calibration.}
    Binary tactile trigger events are shown for calibration trajectories, with FSR responses on top and fingertip taxel responses on bottom. 
    Each vertical mark denotes a frame where at least one valid tactile channel in that modality is active. 
    Black indicates real tactile measurements, orange indicates the nominal uncalibrated simulation, and blue indicates the calibrated simulation. 
    After calibration, the simulated activation pattern better matches real contact onset timing and suppresses spurious activations.
    }
    \label{fig:calibration_alignment}
\end{figure}

Fig.~\ref{fig:calibration_alignment} visualizes tactile trigger events before and after Real2Sim calibration. For each trajectory, we show frames in which at least one valid tactile channel is active, separately for the FSR sensors and fingertip taxels. Compared with the nominal uncalibrated simulator, the calibrated simulator better aligns simulated contact onset with real tactile measurements and suppresses many spurious activations, particularly in the FSR responses. This qualitative result is consistent with the sensing-level metrics reported in Table~\ref{tab:real_calibration}.

\section{Evaluation Protocol and Metrics}
\label{sec:experiment_detail_and_Metrics}

\subsection{Evaluation Protocol}
This section describes the experimental setup used for evaluation.
For both simulation and real-world trials, the object pose is randomized at the start of each grasp attempt. 
In simulation, the object's in-plane position is sampled within a $4\,\mathrm{cm}\times4\,\mathrm{cm}$ region and its 3D orientation is sampled from physically stable resting poses. 
In real-world trials, the object is randomly placed within the same workspace region with a stable resting orientation.
The robot starts from a fixed open-hand pre-grasp pose above the workspace. 
From this state, the policy is expected to establish contact with the object, close the hand to form a grasp, and lift the object using only proprioceptive feedback and binary tactile observations, without visual input, object pose estimates, or object geometry priors.

The success criterion differs between simulation and real-world evaluation. 
In simulation, each episode has a maximum duration of $10\,\mathrm{s}$, including the final holding period. 
A simulated trial is considered successful only if the hand contacts the object, grasps it, lifts it by at least $1.5\,\mathrm{cm}$ above its initial height, and keeps it above this threshold for $1\,\mathrm{s}$ without dropping it. 
In real-world evaluation, we define a task-level success criterion based on physical lifting metrics: a trial is considered successful if the robot successfully lifts the object to a height of at least $3.0\,\mathrm{cm}$ and maintains a stable grasp for over $1\,\mathrm{s}$ without dropping it.

Unless otherwise specified, simulation evaluations use 50 randomized trials per object. 
Since each split contains 10 objects, this corresponds to 500 simulation trials for the seen split and 500 trials for the unseen split. 
For real-world evaluation, we perform 5 hardware trials per object, resulting in 50 trials for seen objects and 50 trials for unseen objects for each method. 
The final real-world success rates are therefore reported as successful trials over 50 trials per split.

\subsection{Tactile Event Fidelity Metrics}
\label{app:tactile_metrics}
Inspired by prior work in tactile sim-to-real transfer~\cite{ding2021simtorealtransferroboticmanipulation}, we use the following metrics to evaluate tactile calibration fidelity in Table~\ref{tab:real_calibration}. 
For each controlled contact trial, we record the real tactile activation 
$y^{\mathrm{real}}_{i,t}\in\{0,1\}$ and the corresponding simulated activation 
$y^{\mathrm{sim}}_{i,t}\in\{0,1\}$ for taxel $i$ at time $t$, using the real tactile stream as reference.

\textbf{Contact Onset Error} $\Delta t_{\mathrm{on}}$ is the absolute time difference between the real and simulated onset contact events, averaged over trials. It measures whether simulated contact appears at the correct time.

\textbf{Activation F1} measures taxel-time agreement between simulated and real tactile events:
\begin{equation}
\mathrm{F1}
=
\frac{2\mathrm{TP}}
{2\mathrm{TP}+\mathrm{FP}+\mathrm{FN}},
\end{equation}
where TP, FP, and FN are counted over all taxel-time activation labels. It measures whether simulation activates the same tactile regions as the real sensor.

\textbf{False Positive Rate (FPR)} measures the fraction of real inactive taxel-time samples that are incorrectly active in simulation:
\begin{equation}
\mathrm{FPR}
=
\frac{\mathrm{FP}}
{\mathrm{FP}+\mathrm{TN}}.
\end{equation}

\section{Additional Ablation Studies on Performance and Design Choices}
\label{app:ablation_experiments}

\subsection{Additional Results on Privileged Tactile Encoder Pretraining}
\label{app:per_object_pretrain}
We provide additional results for privileged tactile encoder pretraining beyond the compact simulation ablation reported in the main manuscript. 
We compare policies with and without the pretrained tactile encoder while keeping the spatial tactile representation, diffusion policy, inputs, and demonstration dataset fixed. 
Thus, the only variable is whether the tactile encoder is pretrained.

Without pretraining, the encoder must learn tactile geometry solely from the imitation objective, which provides weak and indirect supervision for object-level structure. As shown in Table~\ref{tab:sim_pretrain}, privileged pretraining improves average success rates on both seen and unseen objects, indicating that simulator-only geometric and contact labels provide useful supervision for tactile representation learning.

\begin{table*}[h]
\centering
\small
\setlength{\tabcolsep}{5pt}
\renewcommand{\arraystretch}{1.15}
\caption{
Per-object simulation success rate with and without privileged tactile encoder pretraining.
}
\label{tab:sim_pretrain}
\begin{tabular}{lccc | lccc}
\toprule
\multicolumn{4}{c}{\textbf{Seen Objects}}
&
\multicolumn{4}{c}{\textbf{Unseen Objects}}
\\
\cmidrule(lr){1-4}
\cmidrule(lr){5-8}
Object
& w/o Pretrain
& w/ Pretrain
& Improve.
&
Object
& w/o Pretrain
& w/ Pretrain
& Improve.
\\
\midrule

Cube
& 60\% & 62\% & \textcolor{green}{+2\%}
&
Hash-Shape
& 22\% & 42\% & \textcolor{green}{+20\%}
\\

Ball
& 8\% & 56\% & \textcolor{green}{+48\%}
&
C-Shape
& 18\% & 34\% &\textcolor{green}{+16\%}
\\

Box
& 30\% & 48\% & \textcolor{green}{+18\%}
&
E-Shape
& 18\% & 38\% & \textcolor{green}{+20\%}
\\

Cross
& 52\% & 66\% & \textcolor{green}{+14\%}
&
T-Shape
& 28\% & 56\% & \textcolor{green}{+28\%}
\\

CubeBall
& 38\% & 100\% & \textcolor{green}{+62\%}
&
Cylinder
& 2\% & 14\% & \textcolor{green}{+12\%}
\\

Egg
& 16\% & 24\% & \textcolor{green}{+8\%}
&
Fork
& 20\% & 62\% & \textcolor{green}{+42\%}
\\

Big Egg
& 0\% & 8\% & \textcolor{green}{+8\%}
&
Ring
& 50\% & 80\% & \textcolor{green}{+30\%}
\\

H-shape
& 44\% & 76\% & \textcolor{green}{+32\%}
&
Snowman
& 22\% & 48\% & \textcolor{green}{+26\%}
\\

Hollow Cube
& 70\% & 98\% & \textcolor{green}{+28\%}
&
Tetrahedral
& 8\% & 28\% & \textcolor{green}{+20\%}
\\

Hourglass
& 44\% & 66\% & \textcolor{green}{+22\%}
&
Triple
& 12\% & 30\% & \textcolor{green}{+18\%}
\\

\midrule
\textbf{Seen Avg.}
& 36.2\% & 60.4\% & +24.2\%
&
\textbf{Unseen Avg.}
& 20.0\% & 43.2\% & +23.2\%
\\

\bottomrule
\end{tabular}
\end{table*}

We further evaluate the effect of privileged tactile encoder pretraining on the real robot.
To isolate the contribution of pretraining, tactile calibration is disabled in both variants.
As shown in Table~\ref{tab:real_pretrain_ablation_app}, privileged pretraining improves real-world grasp success from 6/100 to 27/100.
These results suggest that the pretrained tactile encoder learns transferable tactile representations that benefit hardware performance even without tactile event calibration.

\begin{table*}[t]
\centering
\small
\setlength{\tabcolsep}{5pt}
\renewcommand{\arraystretch}{1.12}

\begin{minipage}[t]{0.58\textwidth}
\centering
\vspace{0pt}
\captionof{table}{
Real-world performance comparison with and without privileged tactile encoder pretraining.
}
\label{tab:real_pretrain_ablation_app}
\begin{tabularx}{\linewidth}{Xccc}
\toprule
\textbf{Method}
& \textbf{Seen}
& \textbf{Unseen}
& \textbf{Overall} \\
\midrule
w/o Pretrain
& 4/50
& 2/50
& 6/100 \,(6\%) \\
w/ Pretrain
& \textbf{16/50}
& \textbf{11/50}
& \textbf{27/100 \,(27\%)} \\
\bottomrule
\end{tabularx}
\end{minipage}
\hfill
\begin{minipage}[t]{0.38\textwidth}
\centering
\vspace{0pt}
\captionof{table}{
Binary vs. continuous tactile input in simulation on seen objects.
}
\label{tab:tactile_signal_study}
\begin{tabular}{lc}
\toprule
\textbf{Tactile signal} & \textbf{Success} \\
\midrule
Continuous magnitudes & 59.2\% \\
Binary contact events & \textbf{60.4\%} \\
\bottomrule
\end{tabular}
\end{minipage}

\vspace{-4pt}
\end{table*}

\subsection{Binary versus Continuous Tactile Signals}
We compare binary contact events with continuous tactile magnitudes as policy inputs. 
Although continuous signals may encode force intensity, their values are sensitive to sensor mounting, contact geometry, and simulator-specific contact parameters, making calibration and transfer difficult.

Specifically, we created the following continuous variant for comparison. Each tactile channel uses a clipped and per-channel normalized contact magnitude $r_{\tau,i}\in[0,1]$:
\begin{equation}
    s_{\tau,i}^{\mathrm{cont}}
    =
    [p_{\tau,i}^{x},p_{\tau,i}^{y},p_{\tau,i}^{z},r_{\tau,i}],
\end{equation}
where $p_{\tau,i}$ is the forward-kinematics position of tactile site $i$. 
Our binary representation replaces this magnitude with a contact indicator:
\begin{equation}
    s_{\tau,i}^{\mathrm{bin}}
    =
    [p_{\tau,i}^{x},p_{\tau,i}^{y},p_{\tau,i}^{z},a_{\tau,i}],
    \quad a_{\tau,i}\in\{0,1\}.
\end{equation}

For a controlled comparison, both tactile representations are evaluated under the same pretrained setting. 
Each representation uses its own tactile encoder pretrained with the same privileged supervision, and the downstream Diffusion Policy is trained using the same demonstrations and architecture. 
We evaluate both variants on seen objects over 500 simulated grasp trials.
Table~\ref{tab:tactile_signal_study} shows that continuous magnitudes achieve a comparable success rate.
This suggests for blind grasping, the contact occurrence, timing, and location are already sufficient for task execution.
We therefore use binary contacts as a more robust interface to reduce sim-to-real gaps.

\subsection{Ablation on Policy Architecture}
\label{app:policy_ablations}
Our main policy uses Diffusion Policy. To assess this design choice, we compare it with deterministic behavior cloning (BC) and ACT~\cite{zhao2023learningfinegrainedbimanualmanipulation}. All policies are trained on the same simulated expert trajectories and use the same pretrained tactile encoder. As shown in Table~\ref{tab:policy_arch_ablation}, Diffusion Policy matches the best seen-object performance and achieves the highest unseen-object success rate, suggesting that its multimodal action modeling capability improves generalization under partial tactile observability.

\subsection{Ablation on Kinematically Grounded Tactile Inputs}
\label{app:additional_ablations}
We assess whether grounding tactile activations in the hand coordinate frame improves over unstructured binary contact histories. For this purpose, we compare our FK-grounded tactile input against a raw tactile baseline using only the 44-dimensional binary tactile history, without mapping taxels to their forward-kinematics positions. To isolate the representation effect, this ablation excludes privileged encoder pretraining. As shown in Table~\ref{tab:fk_grounding_ablation}, FK grounding improves grasping performance, indicating that spatially locating contact events yields a more enriched tactile representation than an unordered taxel vector.

\begin{table*}[t]
\centering
\small
\setlength{\tabcolsep}{6pt}
\renewcommand{\arraystretch}{1.12}

\begin{minipage}[t]{0.46\textwidth}
\centering
\vspace{0pt}
\captionof{table}{
Ablation on policy architecture.
}
\label{tab:policy_arch_ablation}
\vspace{0.4em}
\begin{tabular}{lccc}
\toprule
\textbf{Policy} 
& \textbf{Seen} 
& \textbf{Unseen}
& \textbf{Avg.} \\
\midrule
BC 
& 59.4\% & 30.2\% & 45.8\% \\
ACT
& 45.0\% & 24.8\% & 34.9\% \\
Diffusion Policy
& \textbf{60.4\%} & \textbf{43.2\%} & \textbf{51.8\%} \\
\bottomrule
\end{tabular}
\end{minipage}
\hfill
\begin{minipage}[t]{0.50\textwidth}
\centering
\vspace{0pt}
\captionof{table}{
Ablation on kinematically grounded tactile inputs.
}
\label{tab:fk_grounding_ablation}
\vspace{0.4em}
\begin{tabular}{lccc}
\toprule
\textbf{Tactile Input} 
& \textbf{Seen} 
& \textbf{Unseen}
& \textbf{Avg.} \\
\midrule
Raw tactile vector
& 33.8\% & 19.8\% & 26.8\% \\
FK-grounded tactile
& \textbf{36.2\%} & \textbf{20.0\%} & \textbf{28.1\%} \\
\bottomrule
\end{tabular}
\end{minipage}

\vspace{-4pt}
\end{table*}

{
\section{Detailed Real-World Failure Mode Analysis}
\label{sec:appendix_failure_mode}

To better understand the remaining bottlenecks of tactile-only blind grasping, we analyze the 73 failed trials among all 100 real-world attempts. 
Each failure is assigned to one of the following modes: \textit{Empty Grasp} (EG), where the hand fails to establish sufficient object contact; \textit{Slip during Lifting} (SL), where the object is initially grasped but drops during lifting; and \textit{Emergency Stop} (ES), where low-level safety limits are triggered.

As shown in Table~\ref{tab:failure_modes}, the most common failure mode is slip during lifting, accounting for 35/73 failures (47.9\%). 
This indicates that the policy often establishes initial contact, but maintaining a stable grasp under real contact dynamics remains challenging. 
Slip failures are especially common for smooth or rounded objects such as balls, cylinders, and other geometries that induce point or line contacts. 
For these objects, small errors in friction, compliance, and contact location can substantially change grasp stability, suggesting that the contact-mechanics sim-to-real gap can dominate over representation or policy differences. 
Our benchmark primarily uses rigid 3D-printed objects, which provide limited compliance and can amplify low-friction slip failures. 
More compliant objects, such as foam or sponge objects, would likely reduce slip, but would also define an easier benchmark with different contact mechanics.

Emergency stops account for 23/73 failures (31.5\%). 
These failures typically occur when sparse tactile feedback and partial observability lead to unsafe arm-hand configurations or workspace-limit violations. 
Empty grasps account for 15/73 failures (20.5\%) and are more common for smooth or hard-to-localize objects such as \textit{Egg}, \textit{Big Egg}, and \textit{Cylinder}, where tactile-only exploration can miss the object before a stable contact is formed.

}
\begin{table*}[t]
\centering
\small
\setlength{\tabcolsep}{6pt}
\renewcommand{\arraystretch}{1.12}
\caption{
Failure mode analysis over 73 failed real-world trials out of 100 attempts.
EG: Empty Grasp; SL: Slip during Lifting; ES: Emergency Stop.
}
\label{tab:failure_modes}
\begin{tabular}{@{}lrrrr@{\hspace{1.2em}}lrrrr@{}}
\toprule
\multicolumn{5}{c}{\textbf{Seen Objects}}
&
\multicolumn{5}{c}{\textbf{Unseen Objects}}
\\
\cmidrule(lr){1-5}
\cmidrule(lr){6-10}
\textbf{Object} & \textbf{EG} & \textbf{SL} & \textbf{ES} & \textbf{Fail.}
&
\textbf{Object} & \textbf{EG} & \textbf{SL} & \textbf{ES} & \textbf{Fail.}
\\
\midrule
Cube        & 0 & 0 & 1 & 1
& Hash-Shape   & 1 & 2 & 1 & 4 \\
Ball        & 0 & 5 & 0 & 5
& C-Shape      & 1 & 2 & 1 & 4 \\
Box         & 2 & 1 & 1 & 4
& E-Shape      & 0 & 1 & 3 & 4 \\
Cross       & 0 & 2 & 2 & 4
& T-Shape      & 0 & 2 & 1 & 3 \\
CubeBall    & 0 & 2 & 2 & 4
& Cylinder     & 0 & 5 & 0 & 5 \\
Egg         & 4 & 0 & 0 & 4
& Fork         & 0 & 0 & 2 & 2 \\
Big Egg     & 4 & 0 & 0 & 4
& Ring         & 0 & 3 & 1 & 4 \\
H-shape     & 1 & 1 & 1 & 3
& Snowman      & 0 & 2 & 2 & 4 \\
Hollow Cube & 0 & 0 & 1 & 1
& Tetrahedral  & 2 & 2 & 0 & 4 \\
Hourglass   & 0 & 0 & 4 & 4
& Triple       & 0 & 5 & 0 & 5 \\
\midrule
\textbf{Seen Total}
& \textbf{11} & \textbf{11} & \textbf{12} & \textbf{34}
&
\textbf{Unseen Total}
& \textbf{4} & \textbf{24} & \textbf{11} & \textbf{39}
\\
\midrule
\multicolumn{10}{c}{
\textbf{Overall:} EG $=15$ (20.5\%), SL $=35$ (47.9\%), ES $=23$ (31.5\%), Total $=73$
}
\\
\bottomrule
\end{tabular}
\end{table*}

\section{Reinforcement Learning Expert Training Details}
\label{app:rl_expert}

To generate successful contact-rich demonstrations for the downstream Diffusion Policy, we train object-specific RL experts with PPO in a parallelized MuJoCo simulation \cite{zakka2025mujocoplayground}. We use an asymmetric actor-critic setup: the critic receives privileged simulator states, while the actor uses only deployable sensory observations. Both networks are MLPs with hidden sizes $[512, 256, 128]$. Training uses $2{,}048$ parallel environments for broad state-space exploration. Hyperparameters are listed in Table~\ref{tab:rl_hyperparams}.

To support sim-to-real transfer, we use an asymmetric actor--critic architecture. The actor observation $\mathcal{O}_t^{\mathrm{actor}}$ contains only hardware-available signals: joint angles, previous joint commands, current and historical binary tactile activations, and proprioceptive tracking errors. The critic state $\mathcal{S}_t^{\mathrm{critic}}$ additionally includes privileged simulator information for value estimation, including noiseless joint positions and velocities, 3D fingertip positions, palm-to-object displacement, object velocities, and external perturbation forces.

We apply task and domain randomization at the start of each episode. The object pose is randomized with planar translation within $\pm 5$ cm and arbitrary 3D rotation, and the hand pre-grasp configuration is perturbed by joint noise within $\pm 0.05$ rad. We also randomize friction, object mass, and actuator dynamics to reduce overfitting to nominal simulation parameters. The full randomization ranges are listed in Table~\ref{tab:domain_randomization}.

\begin{table}[h]
\centering
\small
\caption{Domain Randomization Parameters.}
\label{tab:domain_randomization}
\begin{tabular}{lc}
\toprule
\textbf{Parameter} & \textbf{Randomization Range} \\
\midrule
Fingertip Friction & $\mathcal{U}[0.5, 1.0]$ \\
Object Mass & $[0.8\times, 1.2\times]$ base mass \\
Object Center of Mass Offset & $\pm 5$ mm \\
Hand Joint Friction Loss & $[0.5\times, 2.0\times]$ base value \\
Hand Actuator Gain ($K_p$) & $[0.8\times, 1.2\times]$ base value \\
Hand Joint Damping & $[0.8\times, 1.2\times]$ base value \\
Hand Link Masses & $[0.9\times, 1.1\times]$ base mass \\
\bottomrule
\end{tabular}
\end{table}

\begin{table}[h]
\centering
\small
\caption{Hyperparameters for RL Expert Training (PPO).}
\label{tab:rl_hyperparams}
\begin{tabular}{lc}
\toprule
\textbf{Hyperparameter} & \textbf{Value} \\
\midrule
Algorithm & Proximal Policy Optimization (PPO) \\
Total Timesteps & $200,000,000$ \\
Parallel Environments & $2048$ \\
Batch Size & $2048$ \\
Minibatches & $16$ \\
Epochs (Updates per batch) & $16$ \\
Unroll Length & $40$ for compact objects, $60$ for elongated or rolling objects  \\
Learning Rate & $3 \times 10^{-4}$ \\
Discount Factor ($\gamma$) & $0.99$ \\
Entropy Coefficient & $0.01$ \\
\bottomrule
\end{tabular}
\end{table}

\section{Privileged Tactile Encoder Pretraining Details}
\label{sec:pretrain_details}

We pretrain the tactile encoder using simulator-only privileged supervision to encode object geometry, hand--object contact, and grasp-relevant spatial structure. These privileged labels are used only during pretraining and are unavailable during real-world deployment. During final policy learning, the pretrained tactile encoder is frozen, and only the policy is optimized.

\paragraph{Input representation.}
At each time step $t$, the encoder receives a history window of $H=16$ tactile frames. 
Each frame contains $N_{\mathrm{tac}}=44$ tactile sites. 
For tactile site $i$ at time $\tau$, we concatenate its 3D position from hand forward kinematics and its binary activation:
\begin{equation}
    s_{\tau,i}
    =
    [p_{\tau,i}^{x},p_{\tau,i}^{y},p_{\tau,i}^{z},a_{\tau,i}]
    \in \mathbb{R}^{4}.
\end{equation}
Thus, the tactile history is represented as
\begin{equation}
    \mathcal{S}_t
    =
    \{s_{\tau,i}\}_{\tau=t-H+1:t,\,i=1:N_{\mathrm{tac}}}
    \in \mathbb{R}^{H\times N_{\mathrm{tac}}\times 4}.
\end{equation}
For the first few steps of an episode, we left-pad the history by repeating the earliest available frame.

\paragraph{Tactile encoder.}
We treat each tactile frame as one temporal token. 
The $N_{\mathrm{tac}}\times 4$ features in each frame are flattened and linearly projected to a $d=384$ dimensional embedding. 
After adding learned temporal positional embeddings, the sequence is processed by a Transformer encoder:
\begin{equation}
    Z_t = g_{\psi}(\mathcal{S}_t) \in \mathbb{R}^{H\times d}.
\end{equation}
The encoder uses 6 Transformer layers, 6 attention heads, hidden dimension $384$, and feed-forward dimension $1536$. 
The latent feature of the latest frame, $z_t=Z_t^{H}$, is used as the tactile representation for downstream policy learning.

\paragraph{Privileged decoder.}
During pretraining, we attach a query-based Transformer decoder to the tactile encoder. 
The decoder cross-attends to $Z_t$ and predicts simulator-only privileged quantities:
\begin{align}
    \hat{P}^{\mathrm{hand}}_t &\in \mathbb{R}^{1024\times 3}, &
    \hat{P}^{\mathrm{obj}}_t &\in \mathbb{R}^{1024\times 3}, \\
    \hat{y}^{\mathrm{pose}}_t &\in \mathbb{R}^{7}, &
    \hat{q}^{\mathrm{robot}}_t &\in \mathbb{R}^{22}, &
    \hat{c}_t &\in \mathbb{R}^{N_{\mathrm{tac}}}.
\end{align}
Here $\hat{P}^{\mathrm{hand}}_t$ and $\hat{P}^{\mathrm{obj}}_t$ are hand and object point clouds, $\hat{y}^{\mathrm{pose}}_t$ contains object translation and quaternion, $\hat{q}^{\mathrm{robot}}_t$ is the robot joint configuration, and $\hat{c}_t$ are contact logits for the tactile channels. 
The decoder has the same Transformer width as the encoder and is discarded after pretraining.

\paragraph{Privileged targets.}
All pretraining targets are obtained from simulation. 
We use dense hand and object point clouds, object pose, robot joint configuration, and binary contact labels. 
Let $a_{t,i}\in\{0,1\}$ denote the binary tactile activation of channel $i$ at time $t$, where $a_{t,i}=1$ indicates that the channel is active and $a_{t,i}=0$ indicates no contact. 
Since the tactile observation is binary, the contact label is directly given by the tactile activation,
\begin{equation}
    c_{t,i}=a_{t,i}, \qquad a_{t,i}\in\{0,1\}.
\end{equation}
and we define a frame-level contact mask
\begin{equation}
    m_t=\mathbb{I}\left[\sum_{i=1}^{N_{\mathrm{tac}}} c_{t,i}>0\right].
\end{equation}
Object geometry and pose losses are applied only when $m_t=1$, since tactile observations provide little object information before contact.

\paragraph{Pretraining losses.}
We supervise hand and object geometry using symmetric Chamfer distance:
\begin{equation}
    \mathcal{L}_{\mathrm{pc}}
    =
    \lambda_{\mathrm{hand}}
    \mathrm{CD}(\hat{P}^{\mathrm{hand}}_t,P^{\mathrm{hand}}_t)
    +
    m_t
    \lambda_{\mathrm{obj}}
    \mathrm{CD}(\hat{P}^{\mathrm{obj}}_t,P^{\mathrm{obj}}_t).
\end{equation}
For object pose, we use an $\ell_1$ loss on translation and a sign-invariant quaternion loss on rotation:
\begin{equation}
    \mathcal{L}_{\mathrm{pose}}
    =
    m_t
    \left(
    \|\hat{p}^{\mathrm{obj}}_t-p^{\mathrm{obj}}_t\|_1
    +
    \lambda_{\mathrm{rot}}
    \left(
    1-
    \left|
    \left\langle
    \frac{\hat{q}^{\mathrm{obj}}_t}{\|\hat{q}^{\mathrm{obj}}_t\|_2+\epsilon},
    \frac{q^{\mathrm{obj}}_t}{\|q^{\mathrm{obj}}_t\|_2+\epsilon}
    \right\rangle
    \right|
    \right)
    \right).
\end{equation}
Robot configuration and contact prediction are supervised by
\begin{equation}
    \mathcal{L}_{\mathrm{state}}
    =
    \lambda_{\mathrm{qpos}}
    \|\hat{q}^{\mathrm{robot}}_t-q^{\mathrm{robot}}_t\|_2^2
    +
    \lambda_{\mathrm{contact}}
    \mathrm{FL}(\hat{c}_t,c_t),
\end{equation}
where $\mathrm{FL}$ denotes focal loss, used to handle the sparsity of tactile contacts.

We additionally use two light geometric regularizers on the predicted object point cloud. 
A repulsion loss prevents point collapse, and a hand-object penetration loss discourages predicted object points from lying inside the predicted hand:
\begin{equation}
    \mathcal{L}_{\mathrm{geo}}
    =
    m_t
    \left(
    \lambda_{\mathrm{rep}}\mathcal{L}_{\mathrm{rep}}
    +
    \lambda_{\mathrm{pen}}\mathcal{L}_{\mathrm{pen}}
    \right).
\end{equation}
For frames without contact, we penalize hallucinated object predictions by encouraging the object point cloud and object translation to remain near the origin:
\begin{equation}
    \mathcal{L}_{\mathrm{empty}}
    =
    (1-m_t)
    \left(
    \lambda_{\mathrm{empty\_obj}}
    \frac{1}{N_{\mathrm{obj}}}
    \sum_{\hat{x}\in \hat{P}^{\mathrm{obj}}_t}
    \|\hat{x}\|_2
    +
    \lambda_{\mathrm{empty\_pose}}
    \|\hat{p}^{\mathrm{obj}}_t\|_1
    \right).
\end{equation}

The final pretraining objective is
\begin{equation}
    \mathcal{L}_{\mathrm{pre}}
    =
    \mathcal{L}_{\mathrm{pc}}
    +
    \mathcal{L}_{\mathrm{pose}}
    +
    \mathcal{L}_{\mathrm{state}}
    +
    \mathcal{L}_{\mathrm{geo}}
    +
    \mathcal{L}_{\mathrm{empty}}.
\end{equation}
In our implementation, we use
$\lambda_{\mathrm{hand}}=5.0$,
$\lambda_{\mathrm{obj}}=5.0$,
$\lambda_{\mathrm{rot}}=0.1$,
$\lambda_{\mathrm{qpos}}=1.0$,
$\lambda_{\mathrm{contact}}=2.0$,
$\lambda_{\mathrm{rep}}=0.5$,
$\lambda_{\mathrm{pen}}=0.5$,
$\lambda_{\mathrm{empty\_obj}}=5.0$, and
$\lambda_{\mathrm{empty\_pose}}=2.0$.

We train the encoder and decoder jointly for 200 epochs using AdamW \cite{loshchilov2019decoupledweightdecayregularization} with learning rate $3\times 10^{-4}$. The per-GPU batch size is 256. We use mixed-precision training and clip gradients with maximum norm 1.0. The model is trained with distributed data parallelism across 4 Nvidia A40 GPUs.
After pretraining, the decoder $d_{\omega}$ and all prediction heads are discarded. Only the tactile encoder $g_{\psi}$ is retained and used to compute the tactile feature $z_t$ for policy learning and real-world deployment. No object point cloud, object pose, robot-state label, or simulator contact label is used during deployment.

\begin{table}[h]
\centering
\footnotesize
\setlength{\tabcolsep}{3pt}
\renewcommand{\arraystretch}{1.0}
\caption{Hyperparameters for privileged tactile encoder pretraining.}
\label{tab:pretrain_hyperparams}
\begin{tabular}{lc lc}
\toprule
\textbf{Hyperparameter} & \textbf{Value} 
& \textbf{Hyperparameter} & \textbf{Value} \\
\midrule
History length $H$ & 16 
& Tactile sites $N_{\mathrm{tac}}$ & 44 \\
Per-site input dim. & 4 
& Hidden dimension & 384 \\
Encoder layers / heads & $6/6$ 
& Decoder layers / heads & $6/6$ \\
FFN dimension & 1536 
& Point queries & $64+64$ \\
Points per query & 16 
& Point outputs & $1024+1024$ \\
Pose / robot / contact dim. & $7/22/44$ 
& Optimizer & AdamW \\
Learning rate & $3{\times}10^{-4}$ 
& Epochs & 200 \\
Batch size / GPU & 256 
& Grad. clipping & 1.0 \\
Focal loss $(\alpha,\gamma)$ & $(0.25,2.0)$ 
& Repulsion radius & 0.015 \\
Penetration margin & 0.005 
&  &  \\
\bottomrule
\end{tabular}
\end{table}

{
\section{Diffusion Policy Training and Deployment Details}
\label{app:policy_details}


\paragraph{Diffusion Policy Setup.}
The diffusion policy is conditioned on separate proprioceptive and tactile observations. 
At each timestep, the proprioceptive state is
\[
\mathbf{q}_t=[\mathbf{s}_t,\dot{\mathbf{s}}_t]\in\mathbb{R}^{44},
\]
where $\mathbf{s}_t,\dot{\mathbf{s}}_t\in\mathbb{R}^{22}$ denote joint positions and velocities of the 6-DoF arm and 16-DoF hand. 
The raw tactile observation is
\[
\mathbf{u}_t=[\mathbf{t}_t,\mathbf{f}_t]\in\mathbb{R}^{44},
\]
where $\mathbf{t}_t\in\mathbb{R}^{32}$ and $\mathbf{f}_t\in\mathbb{R}^{12}$ are TwinTac and FSR tactile readings, respectively. 
For kinematically grounded tactile inputs, each tactile site is augmented with its forward-kinematics position, forming
\[
\mathbf{P}_t\in\mathbb{R}^{44\times4},
\]
where each row contains the 3D tactile-site position and the corresponding tactile value. 
For the main policy, this value is a binary contact indicator; for the continuous tactile ablation baseline, it is a normalized contact magnitude.
Unless otherwise specified, policies condition on $T_o=5$ past observations, predict an action chunk of horizon $H=8$, and execute $N_a=3$ actions before replanning. 
Each action $a_t\in\mathbb{R}^{22}$ represents incremental arm and hand joint motion. 
The commanded joint position is computed as
\begin{equation}
    q^{\mathrm{cmd}}_t = \mathrm{clip}(q_t + a_t, q_{\min}, q_{\max}),
\end{equation}
where $q_{\min}$ and $q_{\max}$ are joint limits. 
The commanded positions are tracked by low-level PD controllers. 
We apply the same action scaling and joint-limit constraints in simulation and hardware.

\paragraph{Policy variants.}
The policy variants used in the aforementioned experiments differ in three aspects: tactile input representation, the use of a privileged-pretrained tactile encoder, and the downstream action head. 
They are used in three ablation studies: \textit{Raw-Tactile DP} and \textit{Kinematic-Tactile DP} are compared for kinematically grounded tactile inputs in Section~\ref{app:additional_ablations}; \textit{Kinematic-Tactile DP} and \textit{Pretrained Kinematic-Tactile DP} are compared for privileged tactile encoder pretraining in Section~\ref{app:per_object_pretrain}; and \textit{Pretrained Kinematic-Tactile DP}, \textit{Pretrained Kinematic-Tactile ACT}, and \textit{Pretrained Kinematic-Tactile BC} are compared for policy architecture in Section~\ref{app:policy_ablations}. 
Table~\ref{tab:policy_variants_compact} summarizes the controlled differences among these policy variants.

\begin{table}[h]
\centering
\small
\setlength{\tabcolsep}{4pt}
\renewcommand{\arraystretch}{1.12}
\caption{Policy variants used in different ablation studies.}
\label{tab:policy_variants_compact}
\begin{tabular}{lccc}
\toprule
\textbf{Policy} 
& \textbf{Tactile input} 
& \textbf{Pretrained encoder} 
& \textbf{Action head} \\
\midrule
\textit{Raw-Tactile DP} 
& $\mathbf{u}_t\in\mathbb{R}^{44}$ 
& No 
& DP \\

\textit{Kinematic-Tactile DP} 
& $\mathbf{P}_t\in\mathbb{R}^{44\times4}$ 
& No 
& DP \\

\textit{Pretrained Kinematic-Tactile DP} 
& $\mathbf{P}_t\in\mathbb{R}^{44\times4}$ 
& Yes 
& DP \\

\textit{Pretrained Kinematic-Tactile ACT} 
& $\mathbf{P}_t\in\mathbb{R}^{44\times4}$ 
& Yes 
& ACT \\

\textit{Pretrained Kinematic-Tactile BC} 
& $\mathbf{P}_t\in\mathbb{R}^{44\times4}$ 
& Yes 
& MLP \\
\bottomrule
\end{tabular}
\end{table}


Notably, the tactile representation are different across the variants. 
\textit{Raw-Tactile DP} uses the unordered 44-D tactile vector, \textit{Kinematic-Tactile DP} represents the same activations as $44\times4$ tactile points with spatial coordinates, and \textit{Pretrained Kinematic-Tactile DP} further processes these tactile points with the privileged-pretrained encoder $g_{\psi}$. 
After pretraining, the decoder is discarded and the encoder is frozen for downstream policy learning.

In the full diffusion-policy variant, the low-dimensional observations are encoded by a 3-layer MLP, while the kinematic tactile points are encoded by the frozen tactile encoder into a 384-D feature. 
The two streams are mean-pooled over the observation window and concatenated into a 896-D global condition $\mathbf{c}$. 
The diffusion action head generates $\mathbf{a}\in\mathbb{R}^{H\times22}$ conditioned on $\mathbf{c}$ and is deployed with receding-horizon execution. 
For the policy-head ablation, we keep the same pretrained tactile representation and replace only the downstream action head with ACT or MLP regression. 
The network details of the action heads are summarized in Table~\ref{tab:policy_head_details}.

\begin{table}[h]
\centering
\small
\setlength{\tabcolsep}{5pt}
\renewcommand{\arraystretch}{1.12}
\caption{Network details of the downstream action heads.}
\label{tab:policy_head_details}
\begin{tabular}{lcc}
\toprule
\textbf{Head} & \textbf{Architecture} & \textbf{Objective} \\
\midrule
DP 
& 1D conditional U-Net $[256,512]$, FiLM 
& $\epsilon$-prediction \\

ACT 
& 4-layer Transformer decoder, $d{=}512$, $n_{\text{head}}{=}8$, $d_{\text{ff}}{=}2048$ 
& MSE \\

MLP 
& $896\to2048\to1024\to512\to256\to8{\times}22$ 
& MSE \\
\bottomrule
\end{tabular}
\end{table}

\paragraph{Training details.}
All downstream policies are trained on 10,000 simulated expert trajectories generated as described in Appendix~\ref{app:rl_expert}.
We use the same optimizer, learning-rate schedule, batch size, and evaluation protocol across variants, as listed in Table~\ref{tab:downstream_training_details}. 
Unless otherwise stated, diffusion-policy variants are trained with the standard noise-prediction objective, while ACT and MLP baselines are trained with direct MSE loss on action chunks.}

\makeatletter
\setlength{\@fptop}{0pt}
\setlength{\@fpbot}{0pt plus 1fil}
\makeatother

\begin{table}[!t]
\centering
\small
\setlength{\tabcolsep}{6pt}
\renewcommand{\arraystretch}{1.1}
\caption{Shared downstream training settings.}
\label{tab:downstream_training_details}
\begin{tabular}{lc}
\toprule
\textbf{Hyperparameter} & \textbf{Value} \\
\midrule
Demonstrations & 10,000 simulated trajectories \\
Observation horizon $T_o$ & 5 \\
Prediction horizon $H$ & 8 \\
Executed actions $N_a$ & 3 \\
Action dimension & 22 \\
Optimizer & AdamW \\
Learning rate & $1\times10^{-4}$ \\
AdamW $\beta$ & $(0.95,0.999)$ \\
Weight decay & $1\times10^{-6}$ \\
LR schedule & Cosine decay \\
Warmup steps & 2,000 \\
Batch size & 512 \\
Precision & bfloat16 AMP \\
Evaluation interval & 40k steps \\
\bottomrule
\end{tabular}
\end{table}

\end{document}